\documentclass[letterpaper, 10 pt, conference]{ieeeconf}  

\IEEEoverridecommandlockouts                              

\usepackage{amsmath} 
\usepackage{amssymb}  

\usepackage{url}
\usepackage{multirow}
\usepackage{makecell}
\usepackage{booktabs}
\usepackage{changepage}
\usepackage{balance}
\usepackage[dvipsnames]{xcolor}
\usepackage{graphicx}
\usepackage{tcolorbox}
\usepackage{pifont}
\usepackage[hidelinks]{hyperref}

\newenvironment{codeblock}
  {\begin{tcolorbox}[colback=gray!10, colframe=gray!60, boxrule=0pt, left=1mm, right=1mm, top=0.5mm, bottom=0.5mm]\footnotesize}
  {\end{tcolorbox}}

\newcommand{\figref}[1]{Fig.~\ref{#1}}
\newcommand{\tabref}[1]{Table~\ref{#1}}
\newcommand{\bm}[1]{\mbox{\boldmath{$#1$}}}

\newcommand{\yesmark}{\textcolor{ForestGreen}{\ding{51}}}
\newcommand{\nomark}{\textcolor{BrickRed}{\ding{55}}}
\newcommand{\partialmark}{\textcolor{Orange}{$\triangle$}}

\title{\LARGE \bf
  RoboManipBaselines: A Unified Framework for Imitation Learning in Robotic Manipulation across Real and Simulation Environments
}

\author{
Masaki Murooka$^{1,2}$, Tomohiro Motoda$^{1}$, Ryoichi Nakajo$^{1}$, Hanbit Oh$^{1}$, \\Koshi Makihara$^{1}$, Keisuke Shirai$^{1}$, Tetsuya Ogata$^{1,3,4}$, Yukiyasu Domae$^{1}$
\thanks{$^{1}$Artificial Intelligence Research Center,
National Institute of Advanced Industrial Science and Technology (AIST),
2-3-26 Aomi, Koto-ku, Tokyo 135-0064, Japan.
{\tt\small m-murooka@aist.go.jp}}%
\thanks{$^{2}$CNRS-AIST JRL (Joint Robotics Laboratory), IRL,
1-1-1 Umezono, Tsukuba, Ibaraki 305-8560, Japan.}%
\thanks{$^{3}$Institute for AI and Robotics, Future Robotics Organization, Waseda University,
3-4-1 Okubo, Shinjuku-ku, Tokyo 169-8555, Japan.}%
\thanks{$^{4}$AI Robot Association (AIRoA),
Nagase Hongo Building, 5-24-5 Hongo, Bunkyo-ku, Tokyo 113-0033, Japan.}%
\thanks{This work was supported in part by JST CREST, Japan, Grant Number JPMJCR2553 and JSPS KAKENHI Grant Number 22K17984.}%
}

\begin{document}

\maketitle
\thispagestyle{empty}
\pagestyle{empty}


\begin{abstract}
We present RoboManipBaselines, an open-source software framework for imitation learning research in robotic manipulation.
The framework supports the entire imitation learning pipeline, including data collection, policy training, and rollout, across both simulation and real-world environments.
Its design emphasizes integration through a consistent workflow, generality across diverse environments and robot platforms, extensibility for easily adding new robots, tasks, and policies, and reproducibility through evaluations using publicly available datasets.
RoboManipBaselines systematically implements the core components of imitation learning: environment, dataset, and policy.
Through a unified interface, the framework supports multiple simulators and real robot environments, as well as multimodal sensors and a wide variety of policy models.
We further present benchmark evaluations in both simulation and real-world environments and introduce several research applications, including data augmentation, integration with tactile models, interactive robotic systems, 3D sensing evaluation, and hardware extensions.
These results demonstrate that RoboManipBaselines provides a useful foundation for advancing research and experimental validation in robotic manipulation using imitation learning.
\url{https://isri-aist.github.io/RoboManipBaselines-ProjectPage}
\end{abstract}

\section{Introduction} \label{sec:intro}

In recent years, approaches that generate robot motions through learning from expert demonstration data have been actively studied~\cite{Robomimic:Mandlekar:CoRL2021}.
Furthermore, attempts to apply learning-based motion generation to tasks that were previously difficult to automate with industrial robots are gradually spreading across industrial and service domains worldwide.
Such learning-based approaches require processes such as data collection and model training and execution.
This differs significantly from conventional model-based approaches, which perform motion planning and control based on robot kinematics and dynamics under carefully constructed models of the environment.

Model-based methods have accumulated decades of extensive research, and a wide range of open source robotics software frameworks have been established and released to support needs from academic research to industrial applications~\cite{Pinocchio:Carpentier:SII2019,MoveIt:Chitta:RAM2012,OMPL:Sucan:RAM2012}.
In contrast, although research on learning-based methods has rapidly expanded in recent years and several prominent frameworks have been proposed, the ecosystem of robotics software frameworks that can reliably support diverse research and application needs remains in a fluid state.
Because the processing pipelines required by these approaches differ substantially, it is difficult to directly reuse software assets developed for model-based robotics in learning-based pipelines, which creates a barrier for researchers and practitioners who aim to start robotics learning research or evaluate such methods in real-world applications.


\begin{figure}[tb]
  \centering
  \includegraphics[width=0.99\columnwidth]{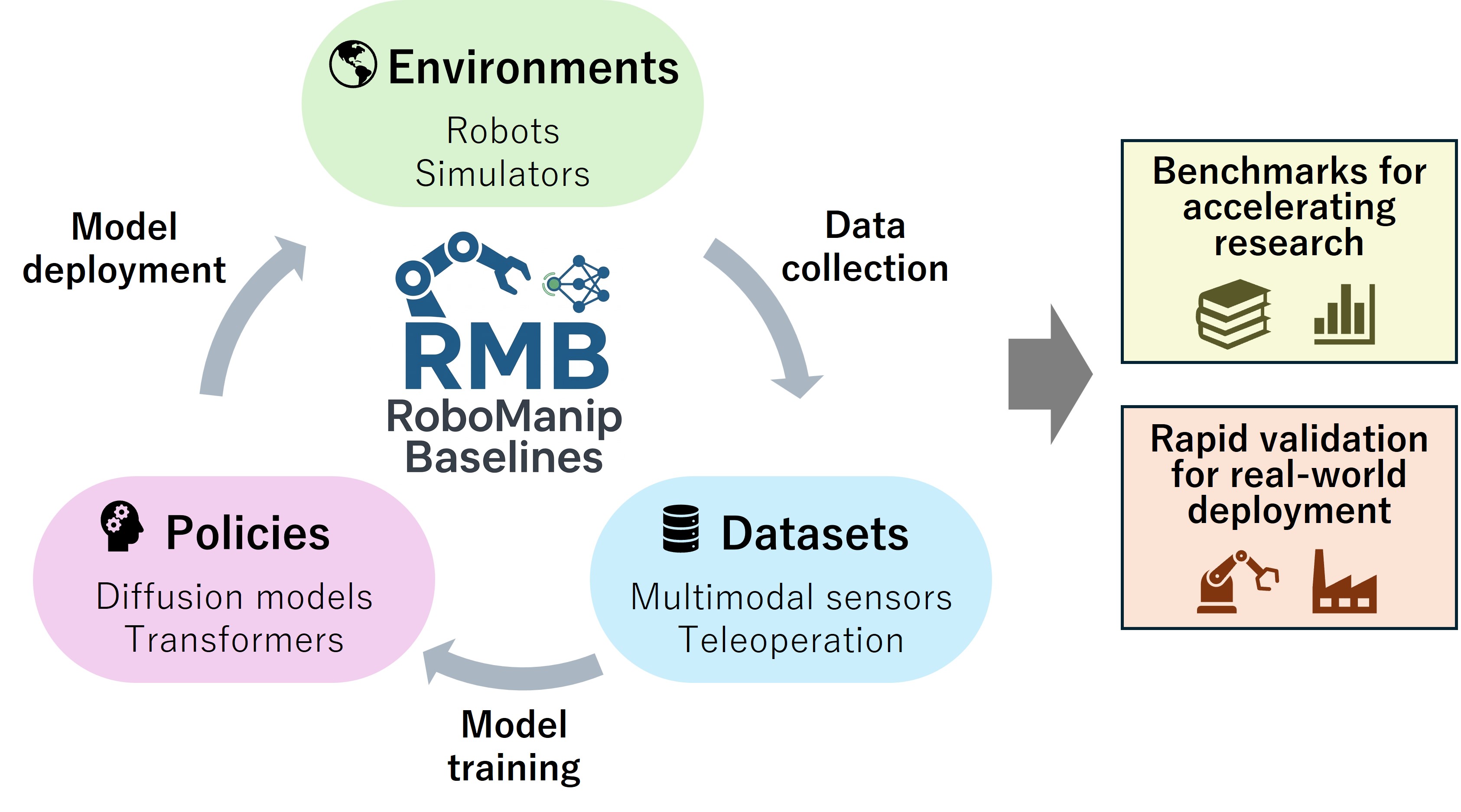}
  \caption{Overview of RoboManipBaselines.}
  \label{fig:overview}
\end{figure}

\begin{table*}[tb]
  \centering
  \renewcommand{\arraystretch}{1.2}
  \caption{Comparison of open frameworks for robot learning.}
  \label{tab:comparison}
  \resizebox{\textwidth}{!}{
    \begin{tabular}{lccccccccc}
      \toprule
      \multirow{2}{*}{\textbf{Framework}} & \multicolumn{3}{c}{\textbf{Environments}} & \multicolumn{2}{c}{\textbf{Embodiments}} & \multicolumn{2}{c}{\textbf{Multimodal Sensors}} & \multicolumn{2}{c}{\textbf{Policy Models}} \\
      \cmidrule(lr){2-4} \cmidrule(lr){5-6} \cmidrule(lr){7-8} \cmidrule(lr){9-10}
      & Real-world & Simulators & \makecell{Deformable\\ objects} & Bimanual & Mobile & Tactile & Point clouds & \makecell{3D} & \makecell{Language-\\conditioned} \\
      \midrule
      robomimic~\cite{Robomimic:Mandlekar:CoRL2021} & \partialmark$^1$ & MuJoCo & \nomark & \yesmark & \partialmark$^2$ & \nomark & \nomark & \nomark & \yesmark \\
      LIBERO~\cite{LIBERO:Liu:NeurIPS2023} & \nomark & MuJoCo & \nomark & \nomark & \nomark & \nomark & \nomark & \nomark & \yesmark \\
      ManiSkill~\cite{Maniskill:Tao:RSS2025} & \nomark & SAPIEN & \nomark & \yesmark & \yesmark & \nomark & \yesmark & \nomark & \nomark \\
      RoboHive~\cite{RoboHive:Kumar:NeurIPS2023} & \yesmark & MuJoCo & \yesmark & \yesmark & \yesmark & \nomark & \yesmark & \nomark & \nomark \\
      RoboCasa~\cite{RoboCasa:Nasiriany:RSS2024} & \partialmark$^1$ & MuJoCo & \nomark & \yesmark & \yesmark & \nomark & \yesmark & \nomark & \yesmark \\
      D3IL~\cite{D3IL:Jia:ICLR2024} & \nomark & MuJoCo & \nomark & \nomark & \nomark & \nomark & \nomark & \nomark & \nomark \\
      COLOSSEUM~\cite{Colosseum:Pumacay:RSS2024} & \yesmark & CoppeliaSim & \yesmark & \nomark & \nomark & \nomark & \yesmark & \yesmark & \yesmark \\
      LeRobot~\cite{LeRobot:Cadene:ICLR2026} & \yesmark & MuJoCo & \nomark & \yesmark & \yesmark & \nomark & \nomark & \nomark & \yesmark \\
      RoboVerse~\cite{RoboVerse:Geng:RSS2025} & \partialmark$^3$ & 7 simulators & \yesmark & \yesmark & \yesmark & \yesmark & \yesmark & \nomark & \yesmark \\
      \textbf{RoboManipBaselines} & \yesmark & 3 simulators & \yesmark & \yesmark & \yesmark & \yesmark & \yesmark & \yesmark & \yesmark \\
      \bottomrule
    \end{tabular}
  }
  \\
  \vspace{3mm}
  \begin{adjustwidth}{10pt}{0pt}
    \raggedright
    $^1$ While the framework has been applied to real robots in some reports, it does not fully support the end-to-end workflow from data collection to model deployment across multiple robot types.\\
    $^2$ An external dataset for mobile manipulators, MOMART~\cite{MOMART:Wong:CoRL2022}, has been released in a format compatible with robomimic. However, mobile manipulators are not directly supported within robomimic itself.\\
    $^3$ RoboVerse provides sim-to-real deployment and real-to-sim reconstruction, but teleoperation-based data collection is primarily described for simulation environments and not explicitly demonstrated in real-world settings.
  \end{adjustwidth}
\end{table*}

In this paper, we present \textbf{RoboManipBaselines}, an open source general purpose software framework for imitation learning in robot manipulation.
The framework is developed to address the challenges discussed above, and we describe its design philosophy and application examples.
As illustrated in \figref{fig:overview}, RoboManipBaselines supports the entire imitation learning cycle, including environment, dataset, and policy.
To facilitate both research and practical experimentation, the framework is designed based on the following four principles.

\begin{itemize}
\item \textbf{Integration}: providing an end-to-end workflow from data collection to policy training and rollout
\item \textbf{Generality}: supporting diverse environments in both simulation and the real world
\item \textbf{Extensibility}: allowing new robots, tasks, and policy models to be added with minimal effort
\item \textbf{Reproducibility}: enabling systematic benchmarking with publicly available datasets
\end{itemize}

To support these principles in practice, RoboManipBaselines currently provides over 10 simulated robot environments across 3 simulators, 2 real robot environments, publicly available datasets with over 1,500 episodes across 12 tasks, and 9 policy implementations.
The diverse application examples presented in this paper demonstrate that RoboManipBaselines facilitates research in robot imitation learning from multiple perspectives, and we believe that it provides a useful software framework for the robot learning research community.


The remainder of this paper is organized as follows.
Section~II reviews related work, and Section~III describes the imitation learning problem formulation.
Section~IV presents the software design, and Section~V describes how to use the framework.
Section~VI reports benchmark results, and Section~VII introduces diverse application examples.
Finally, Section~VIII discusses the limitations, and Section~IX concludes the paper and outlines future directions.


\section{Related Works} \label{sec:related-works}

\subsection{Imitation Learning for Robotic Manipulation}

Imitation learning has been actively studied in recent years as a method for acquiring advanced manipulation skills from expert demonstration data without requiring detailed environment modeling or manually designed control rules~\cite{Robomimic:Mandlekar:CoRL2021,ImitationSurvey:Fang:IJIRA2019,LfDSurvey:Argall:RAS2009}.
Recent research has introduced various improvements to visuomotor policies represented by neural networks, such as future image prediction~\cite{DoorOpen:Ito:SR2022}, action chunking~\cite{ALOHA:Zhao:RSS2023}, and the use of diffusion models and flow matching~\cite{DiffusionPolicy:Chi:IJRR2024,ManiFlow:Yan:CoRL2025}.
These developments have improved robustness to disturbances, enabled the handling of multimodal action distributions, and increased inference speed, making it possible to address dexterous manipulation and long horizon tasks~\cite{VLASurvey:Motoda:Techrxiv2025}.
In addition to camera images, many policy models have been proposed that leverage diverse sensing modalities, including 3D point clouds~\cite{ManiFlow:Yan:CoRL2025,3DDiffusionPolicy:Ze:RSS2024}, tactile measurements~\cite{UniT:Xu:RAL2025,TACT:Murooka:RAL2025}, and audio signals~\cite{AudioIL:Wang:CoRL2025}.
Furthermore, by training on large scale datasets, recent models have demonstrated the ability to follow language instructions and perform multiple tasks within a single policy~\cite{MtAct:Bharadhwaj:ICRA2024,pi0:Black:arXiv2024,GR00T:Bjorck:arXiv2025}.
Beyond policy models, active research has also explored systems that include novel hardware for collecting training data~\cite{GELLO:Wu:IROS2024,UMI:Chi:RSS2024} as well as methods for constructing large scale datasets~\cite{RLDS:Ramos:arXiv2021}.


\subsection{Open-source Frameworks and Benchmarks}

For model-based approaches, many reliable software frameworks have been released based on decades of research, covering topics from robot kinematics and dynamics~\cite{Pinocchio:Carpentier:SII2019} to motion planning~\cite{MoveIt:Chitta:RAM2012,OMPL:Sucan:RAM2012}.
In contrast, although several software frameworks for learning-based approaches have been proposed and released in recent years, the ecosystem is still evolving compared with that of model-based methods.

Table~\ref{tab:comparison} summarizes representative software frameworks~\cite{Robomimic:Mandlekar:CoRL2021,LIBERO:Liu:NeurIPS2023,Maniskill:Tao:RSS2025,RoboHive:Kumar:NeurIPS2023,RoboCasa:Nasiriany:RSS2024,D3IL:Jia:ICLR2024,Colosseum:Pumacay:RSS2024,LeRobot:Cadene:ICLR2026,RoboVerse:Geng:RSS2025} that consistently support the main components of the imitation learning cycle shown in \figref{fig:overview}.
Although the comparison is not exhaustive, it highlights several strengths of RoboManipBaselines, including seamless support for both simulation and real-world environments, compatibility with multiple simulators, support for diverse robot embodiments, and policy models based on multimodal sensing.

In addition, several benchmarks and datasets have been released to support imitation learning research.
As a benchmark, RLBench~\cite{RLBench:James:RAL2020} was initially proposed as a manipulation benchmark in simulation, and more recently RoboArena~\cite{RoboArena:Atreya:CoRL2025} has been introduced to enable unified evaluation of policies on real robots contributed by researchers worldwide.
For datasets, large-scale collections such as DROID~\cite{DROID:Khazatsky:RSS2024}, which provides demonstrations in standardized manipulation environments, and Open X-Embodiment~\cite{OpenX:OpenX:ICRA2024}, which aggregates diverse datasets contributed by research groups around the world, have been released.
RoboManipBaselines is designed with sufficient flexibility to integrate with and leverage these resources.




\section{Problem Formulation}

We describe the problem formulation of imitation learning for robotic manipulation addressed by RoboManipBaselines.
An environment $\mathcal{E}$ is defined by a state space $\mathcal{S}$, an action space $\mathcal{A}$, and a state transition function $f : \mathcal{S} \times \mathcal{A} \to \mathcal{S}$.
At time step $t$, the robot observes a state $s_t \in \mathcal{S}$ and selects an action $a_t \in \mathcal{A}$, resulting in the next state $s_{t+1} = f(s_t, a_t)$.

In imitation learning, trajectories of robot observations and actions $\tau = \{(s_t, a_t)\}_{t=1}^{T}$ are first collected from expert demonstrations.
From these trajectories, a dataset $\mathcal{D} = \{\tau^{(i)}\}_{i=1}^{N}$ is constructed.
Based on this dataset, a policy $\pi : \mathcal{S} \to \mathcal{A}$ that maps states to actions is learned.
The policy $\pi$ is typically parameterized by a neural network and trained using supervised learning methods such as behavior cloning.



\begin{figure}[tb]
  \centering
  \includegraphics[width=1.0\columnwidth]{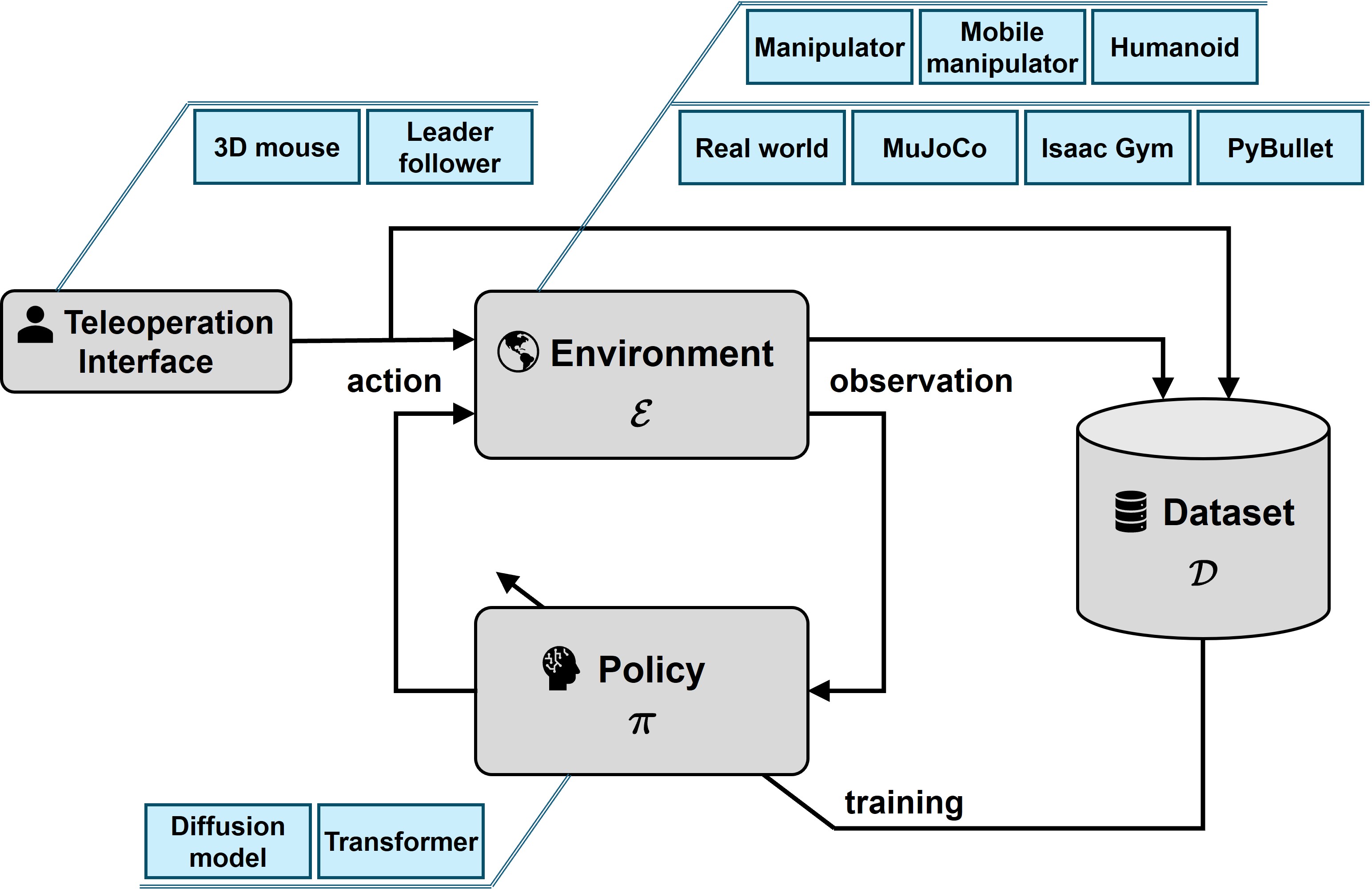}
  \caption{Components of imitation learning.}
  \label{fig:system}
\end{figure}

\section{Framework Design}

\begin{figure*}[tb]
  \centering
  \includegraphics[width=0.19\textwidth]{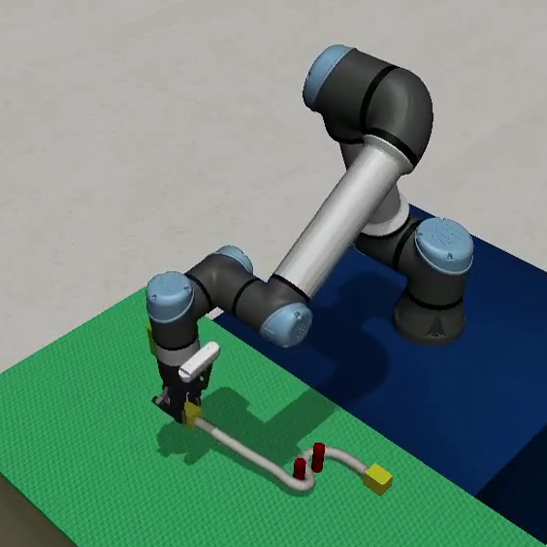}
  \includegraphics[width=0.19\textwidth]{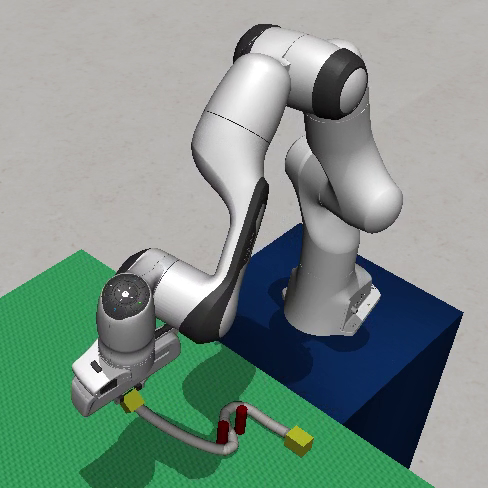}
  \includegraphics[width=0.19\textwidth]{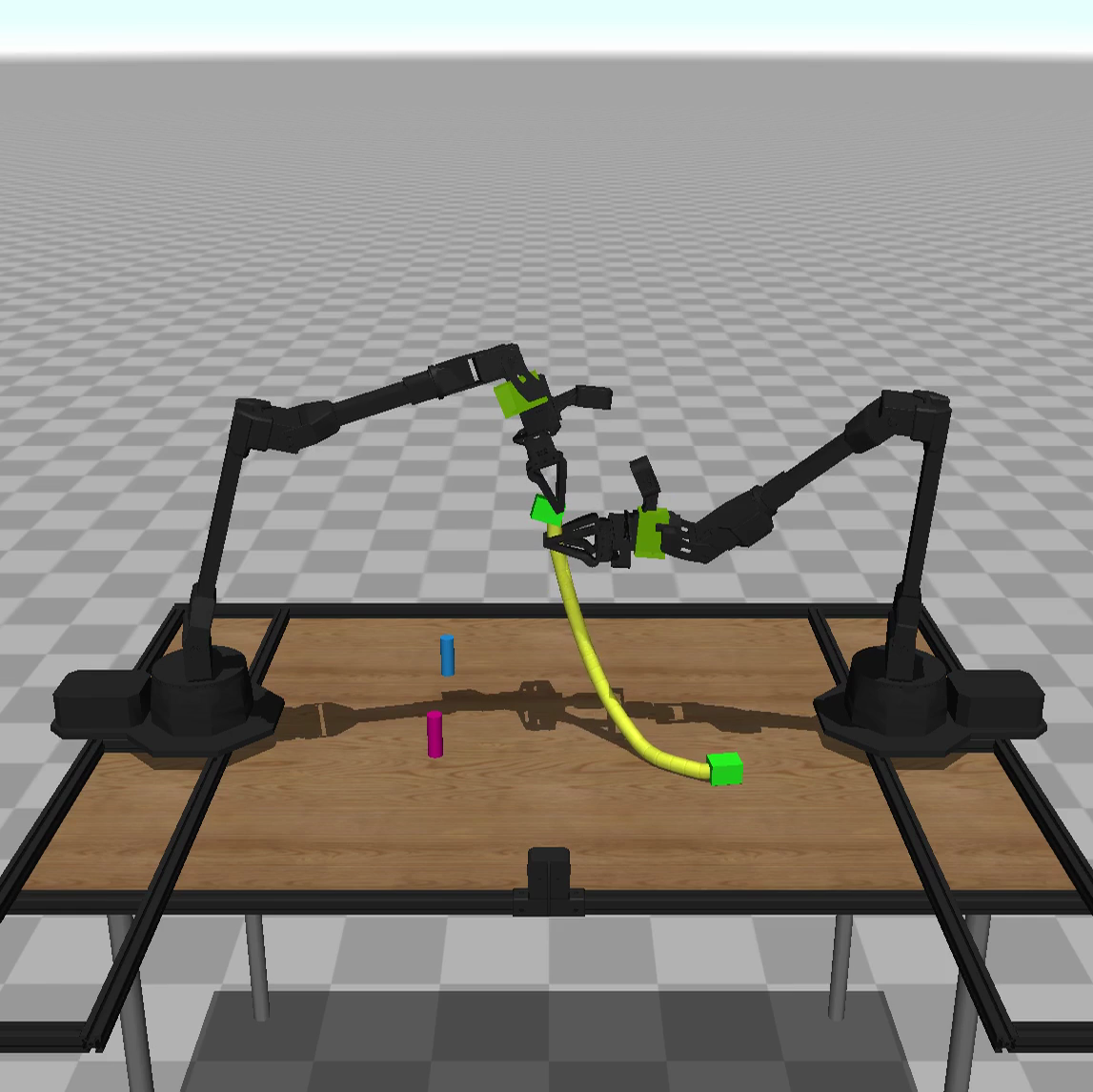}
  \includegraphics[width=0.19\textwidth]{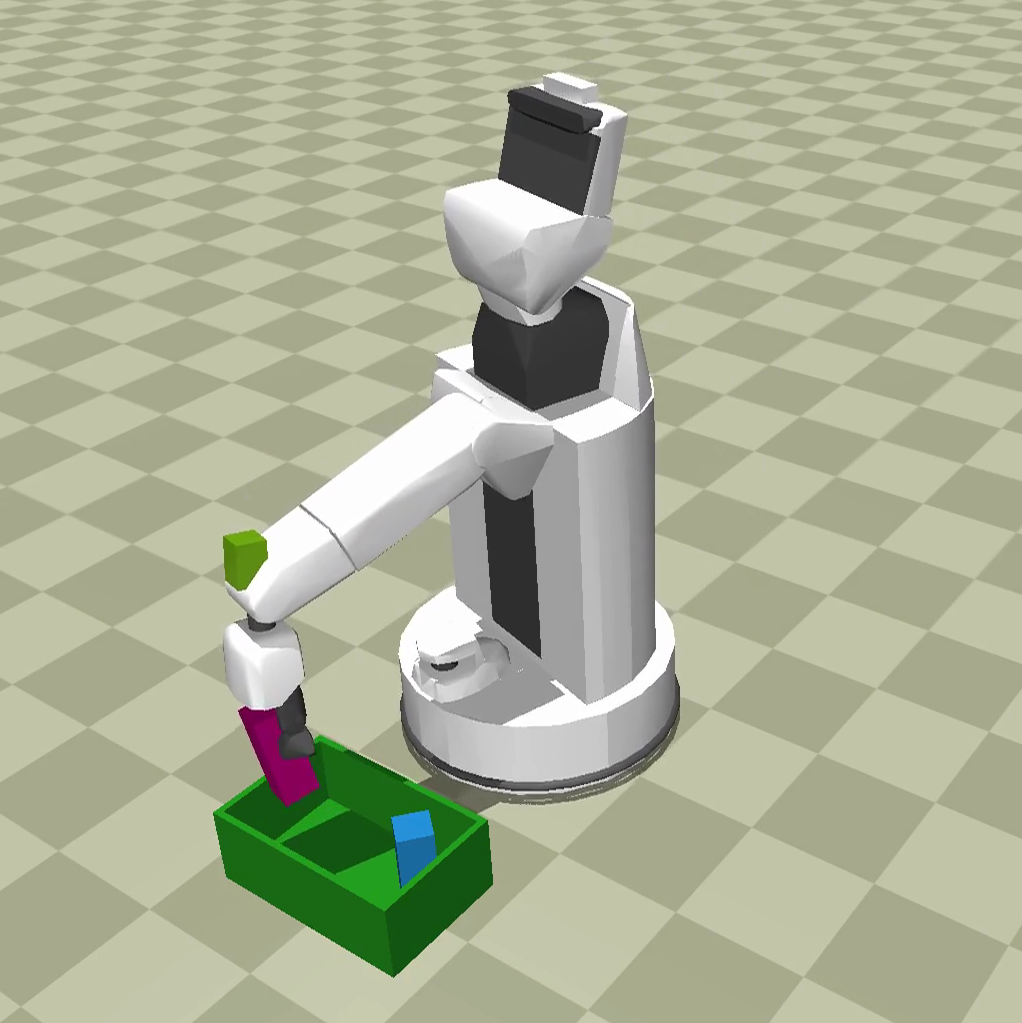}
  \includegraphics[width=0.19\textwidth]{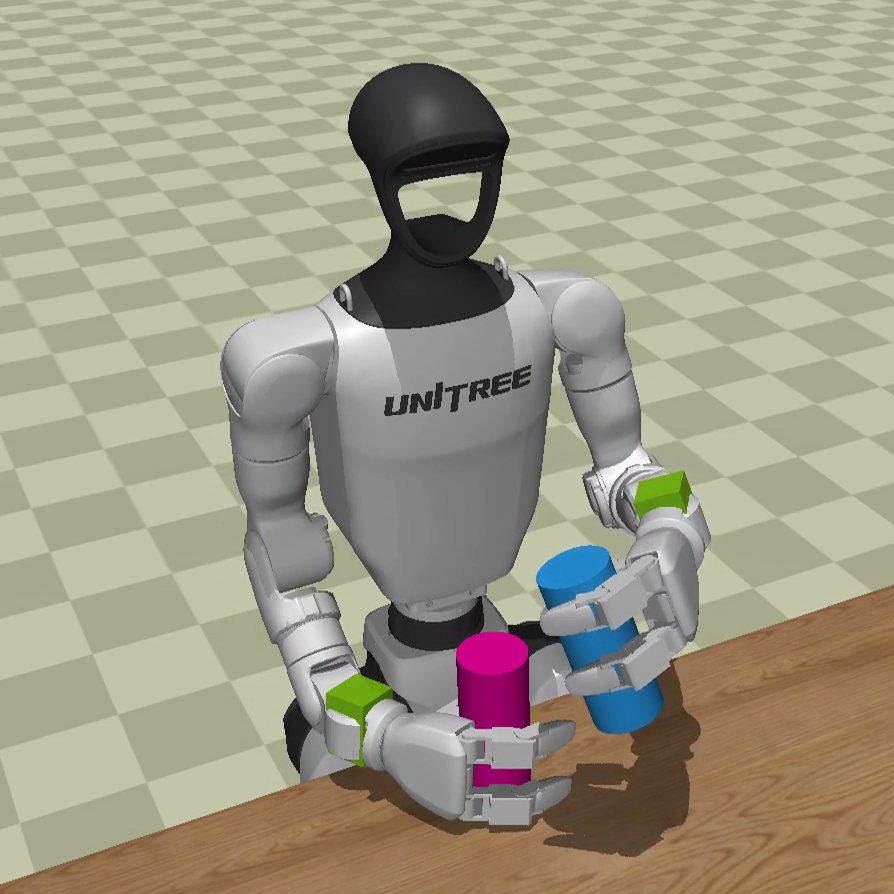}\\
  \begin{minipage}{0.19\textwidth}
    \begin{center} \footnotesize UR5e in MuJoCo \end{center}
  \end{minipage}
  \begin{minipage}{0.19\textwidth}
    \begin{center} \footnotesize FR3 in MuJoCo \end{center}
  \end{minipage}
  \begin{minipage}{0.19\textwidth}
    \begin{center} \footnotesize ALOHA in MuJoCo \end{center}
  \end{minipage}
  \begin{minipage}{0.19\textwidth}
    \begin{center} \footnotesize HSR in MuJoCo \end{center}
  \end{minipage}
  \begin{minipage}{0.19\textwidth}
    \begin{center} \footnotesize G1 in MuJoCo \end{center}
  \end{minipage}\\
  \vspace{3mm}
  \includegraphics[width=0.19\textwidth]{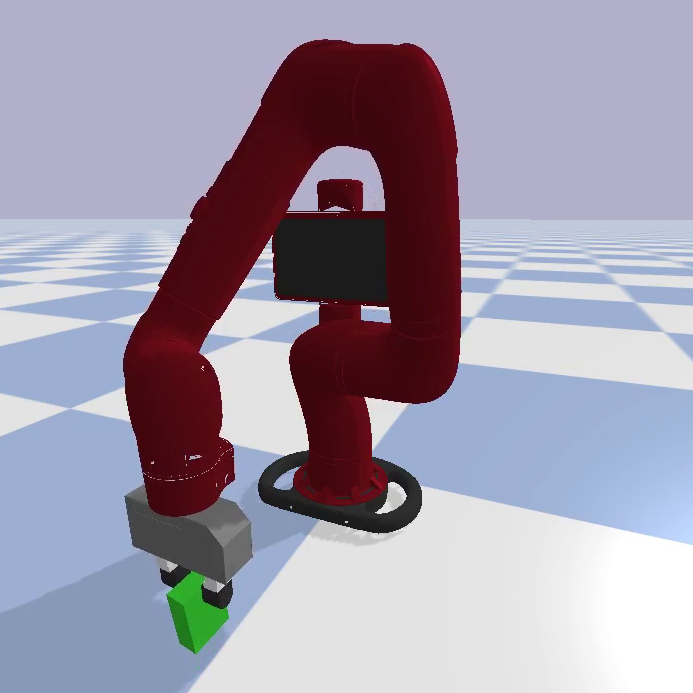}
  \includegraphics[width=0.19\textwidth]{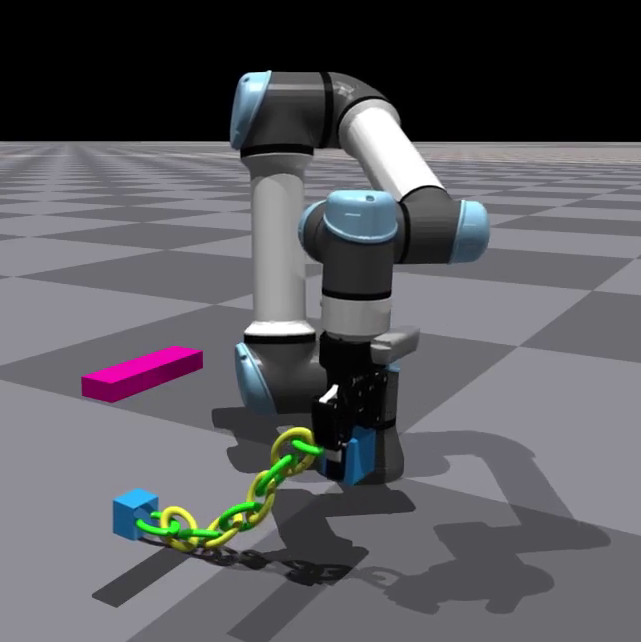}
  \includegraphics[width=0.19\textwidth]{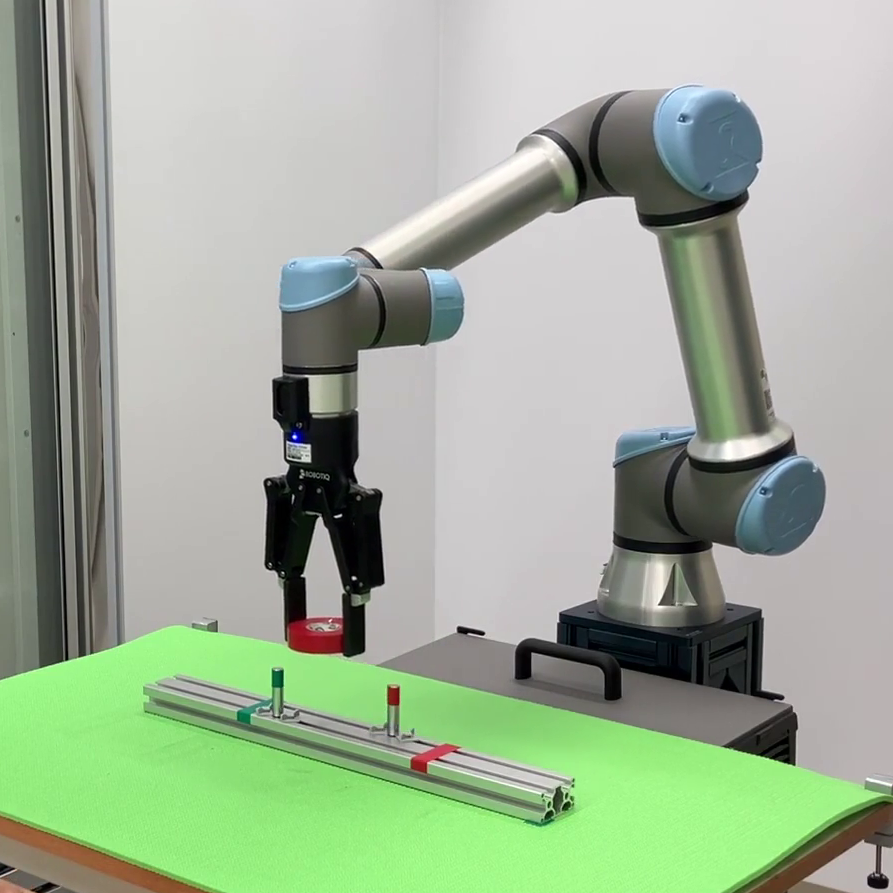}
  \includegraphics[width=0.19\textwidth]{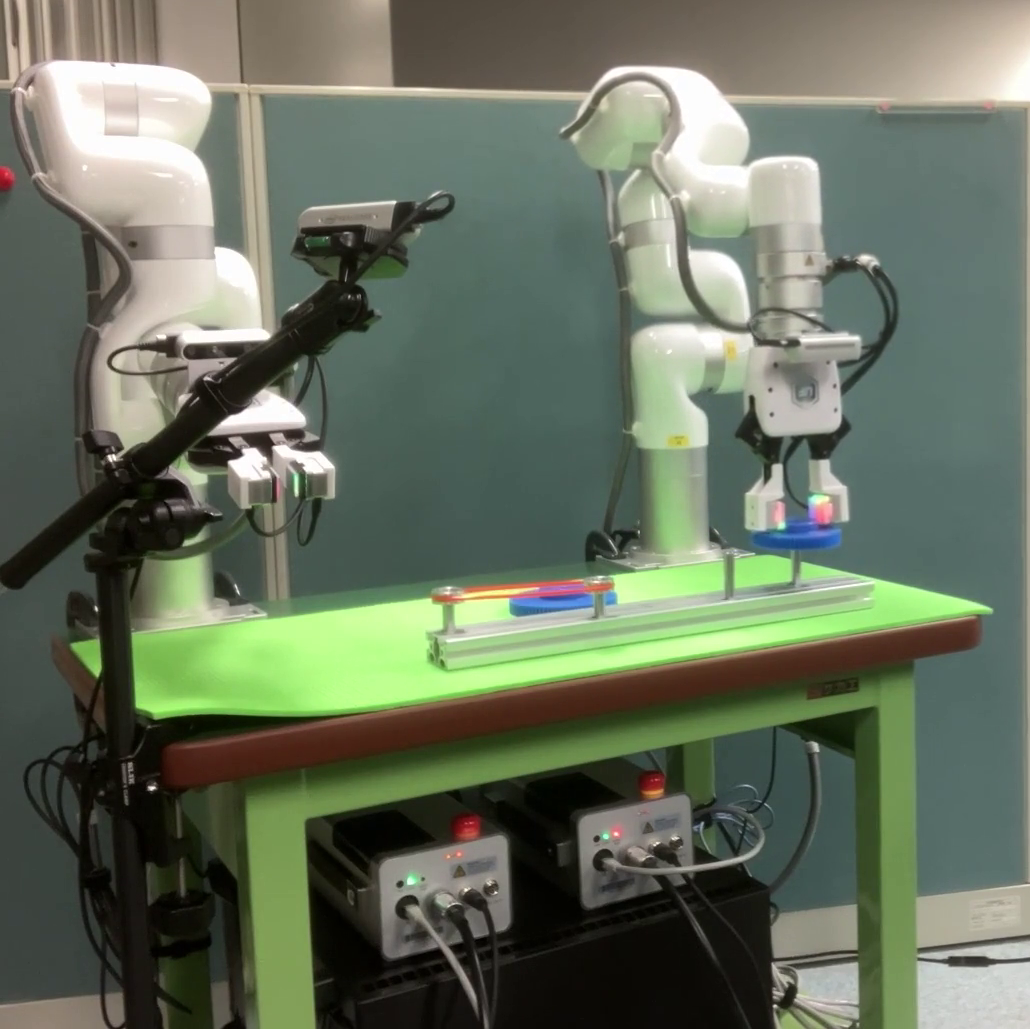}
  \includegraphics[width=0.19\textwidth]{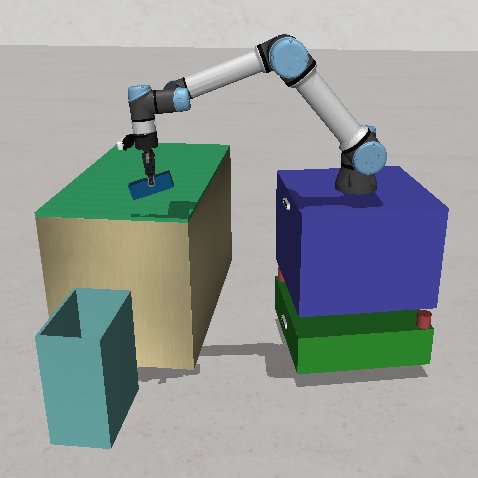}\\
  \begin{minipage}{0.19\textwidth}
    \begin{center} \footnotesize Sawyer in PyBullet \end{center}
  \end{minipage}
  \begin{minipage}{0.19\textwidth}
    \begin{center} \footnotesize UR5e in Isaac Gym \end{center}
  \end{minipage}
  \begin{minipage}{0.19\textwidth}
    \begin{center} \footnotesize UR5e real robot \end{center}
  \end{minipage}
  \begin{minipage}{0.19\textwidth}
    \begin{center} \footnotesize xArm7 real robot \end{center}
  \end{minipage}
  \begin{minipage}{0.19\textwidth}
    \begin{center} \footnotesize Original mobile manipulator \end{center}
  \end{minipage}
  \caption{Simulation and real environments.}
  \label{fig:env}
\end{figure*}

The main components of imitation learning are the environment $\mathcal{E}$, dataset $\mathcal{D}$, and policy $\pi$, as illustrated in \figref{fig:system}.
For each component, we describe how RoboManipBaselines is designed to satisfy the principles of integration, generality, extensibility, and reproducibility.


\subsection{Environment}

\begin{figure}[tb]
  \centering
  \includegraphics[width=0.48\columnwidth]{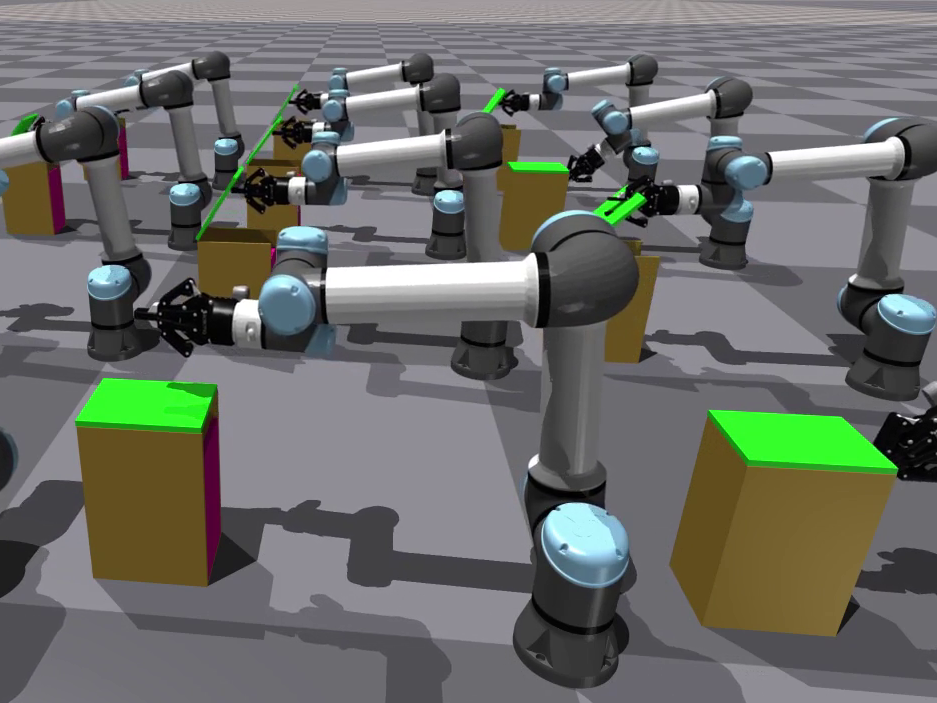}
  \includegraphics[width=0.48\columnwidth]{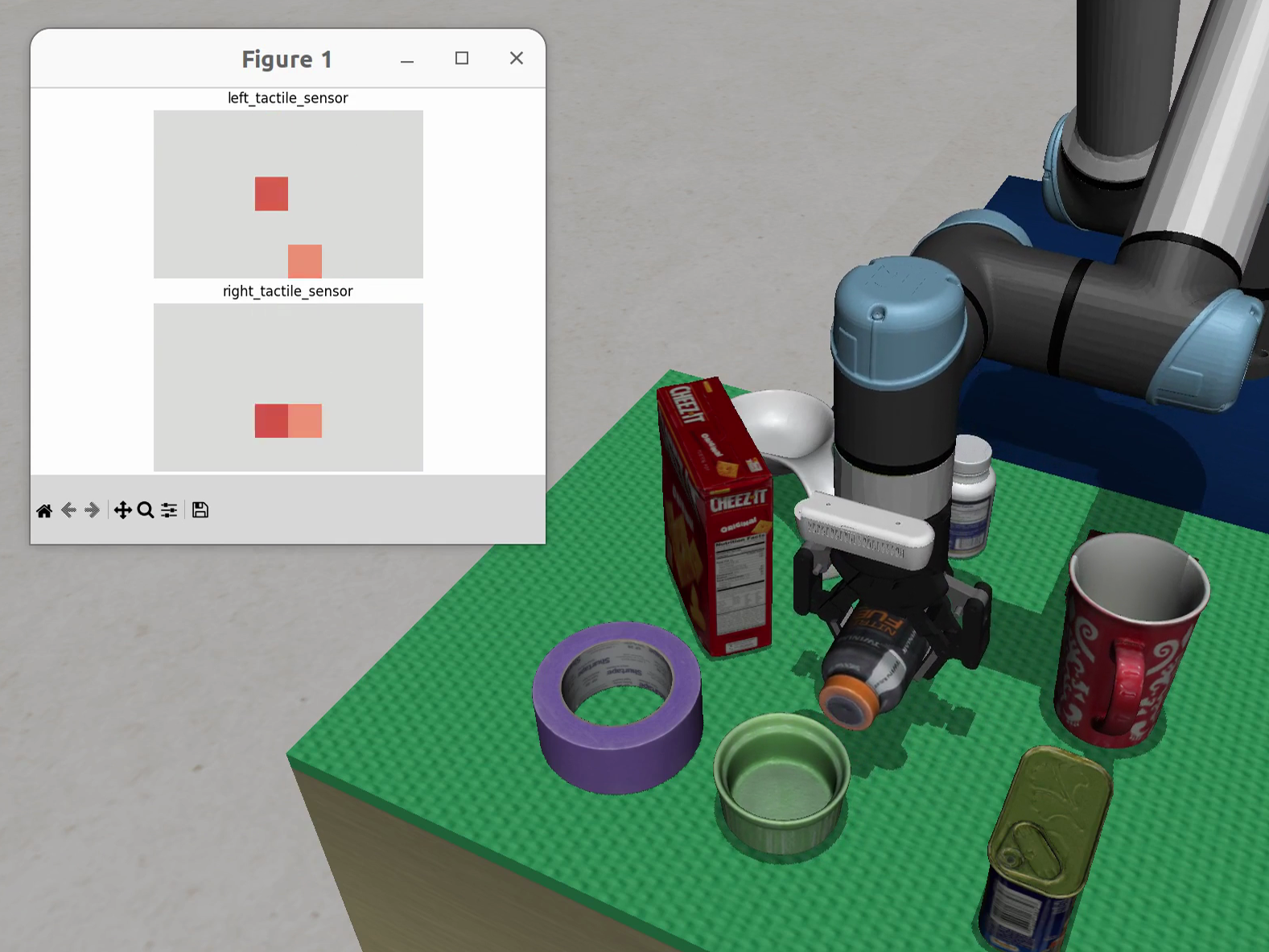}\\
  \begin{minipage}{0.48\columnwidth}
    \begin{center} \footnotesize Parallel simulation \end{center}
  \end{minipage}
  \begin{minipage}{0.48\columnwidth}
    \begin{center} \footnotesize Tactile sensor plugin \end{center}
  \end{minipage}
  \caption{Simulation environments with advanced features.}
  \label{fig:env2}
\end{figure}

In RoboManipBaselines, various environments, including those illustrated in \figref{fig:env}, are implemented using an interface compatible with OpenAI Gym~\cite{OpenAIGym:Brockman:arXiv2016}, which enables seamless switching and unified handling of different environments.
By adopting a design that treats simulation and real-world environments in a consistent manner, programs initially developed and validated in simulation can be directly deployed on real robots, and data collected from both simulation and real environments can be combined for training and evaluation.
Furthermore, when introducing a new simulator or robot, developers can easily extend the framework by inheriting the abstract environment class and overriding a small number of methods.

As simulation backends, RoboManipBaselines currently supports three simulators: MuJoCo~\cite{MuJoCo:Todorov:IROS2012}, Isaac~Gym~\cite{IsaacGym:Makoviychuk:NeurIPS2021}, and PyBullet~\cite{PyBullet:Coumans:URL2016}\footnote{A comprehensive list of available environments can be found at \\\scriptsize{\url{https://github.com/isri-aist/RoboManipBaselines/blob/master/doc/environment_catalog.md}}}.
Within MuJoCo, a wide range of robots are available, including manipulators (Universal Robots UR5e, UFACTORY xArm~7, Franka Research~3, Kinova~Gen3, and FANUC CRX-5iA), a dual-arm manipulator (ALOHA), a mobile manipulator (TOYOTA HSR), and a humanoid robot (Unitree G1\footnote{Unitree G1 is a biped humanoid robot with two arms. In RoboManipBaselines, the leg joints are currently fixed and only the upper body is used for manipulation.}).
For real-world environments, UR5e and xArm~7 are currently supported, and support for Franka Research~3 and Kinova~Gen3 is planned in the near future.

By inheriting and extending the open source code of RoboManipBaselines, users can implement only the minimal robot-specific components as closed source code without copying or modifying the original framework.
This design makes it straightforward to clearly separate public and private code.
Such a modular structure allows developers to continuously incorporate updates from the RoboManipBaselines repository while maintaining custom implementations for their own robots.

For manipulation tasks in simulation, the framework provides more than ten tasks, including those involving deformable objects such as cables, cloth, and granular materials.
Simulation environments also enable the use of advanced features provided by the simulators.
\figref{fig:env2} shows two examples of such extended simulation environments.
In one example, the parallel simulation capability of Isaac~Gym is used to send teleoperation commands to multiple robots simultaneously, enabling parallelized data collection.
In another example, a MuJoCo tactile simulation plugin is imported to enable grid-based tactile sensing at the fingertips of the gripper.





\subsection{Dataset}

RoboManipBaselines publicly releases collected datasets to facilitate research and to serve as benchmarks for imitation learning.
In particular, simulation datasets are emphasized because, once the software environment is set up, anyone can reproduce model training and rollout using the released datasets, enabling rapid comparison of new methods\footnote{A comprehensive list of available datasets can be found at \\\scriptsize{\url{https://github.com/isri-aist/RoboManipBaselines/blob/master/doc/dataset_list.md}}}.
As representative data, we provide a dataset consisting of 100 episodes for each of 10 single-task manipulation scenarios, totaling 1000 episodes.
In addition, for language-conditioned tasks, we release a dataset for a Pick\&Place task in which nine types of objects are placed into two types of baskets.
For the 18 combinations of objects and baskets, 30 episodes are collected for each combination, resulting in a total of 540 episodes.
The robot trajectories contained in these datasets are smooth and efficient motions generated by an experienced operator using a 3D mouse.

\begin{figure}[tb]
  \centering
  \includegraphics[width=0.45\columnwidth]{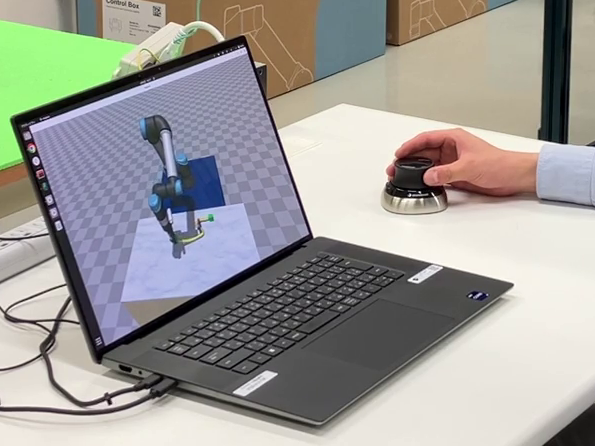}
  \hspace{1mm}
  \includegraphics[width=0.45\columnwidth]{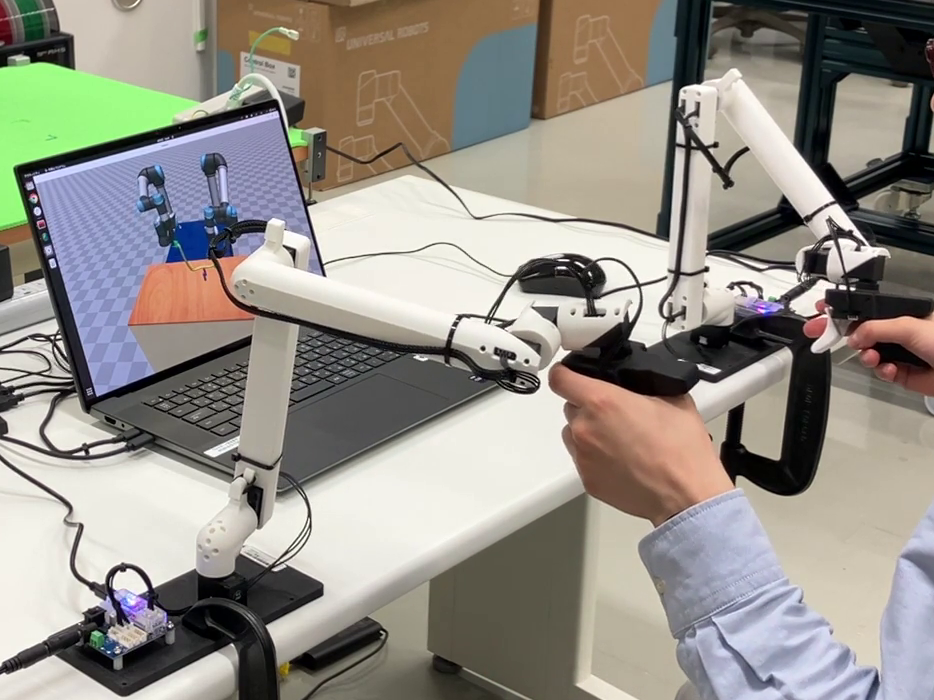}\\
  \vspace{1mm}
  \begin{minipage}{0.45\columnwidth}
    \begin{center} \footnotesize 3D mouse \end{center}
  \end{minipage}
  \hspace{1mm}
  \begin{minipage}{0.45\columnwidth}
    \begin{center} \footnotesize Leader-follower system\\ GELLO~\cite{GELLO:Wu:IROS2024} \end{center}
  \end{minipage}
  \caption{Teleoperation interface.}
  \label{fig:teleop}
\end{figure}

Users can also easily collect additional datasets.
As shown in \figref{fig:teleop}, multiple teleoperation interfaces are supported, including a keyboard, a 3D mouse, and the leader-follower system GELLO~\cite{GELLO:Wu:IROS2024}.
The framework is also designed to allow new interfaces, such as VR-based systems, to be integrated with minimal effort.



\begin{figure}[tb]
  \centering
  \includegraphics[width=0.8\columnwidth]{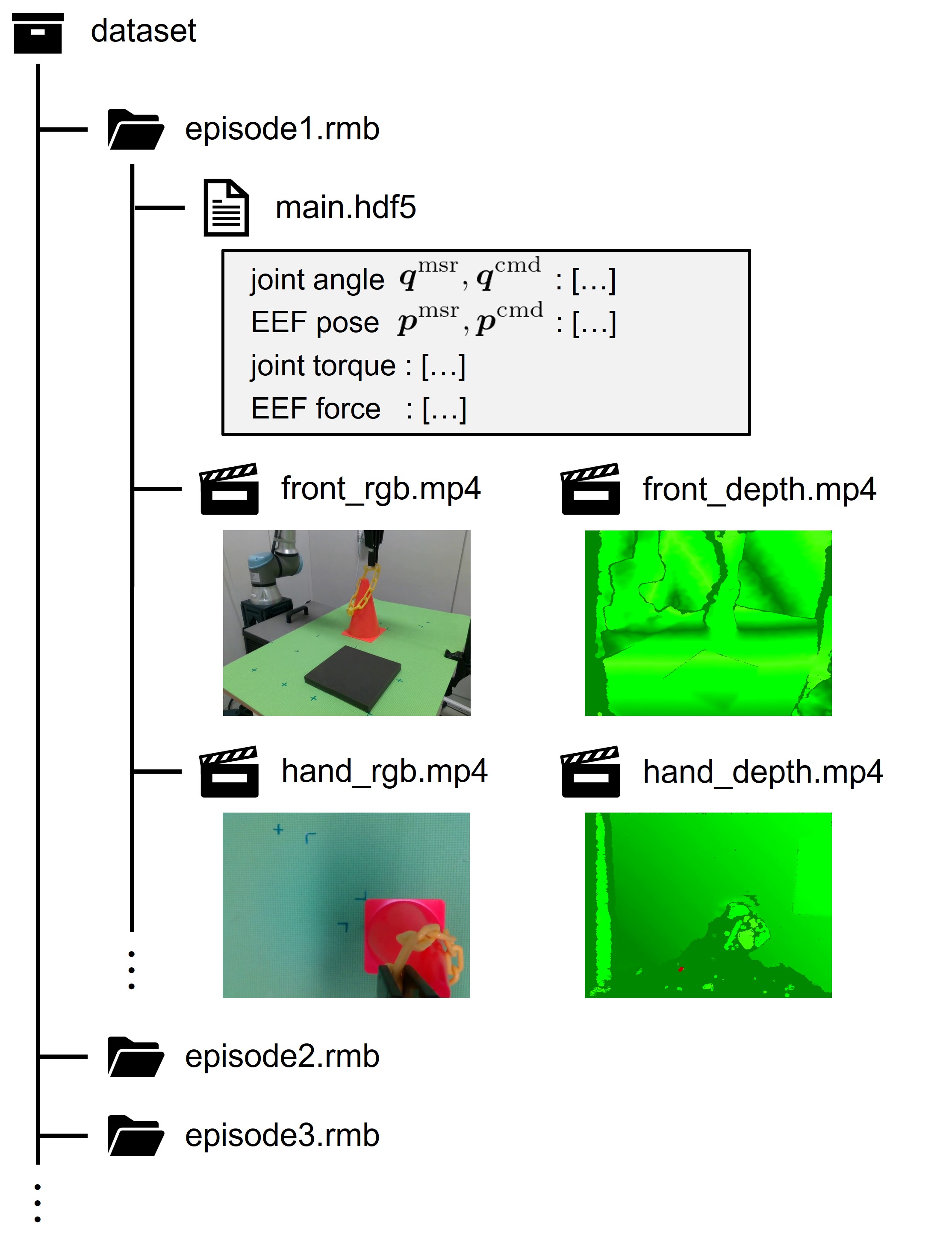}
  \caption{
    Data structure.\newline
    \footnotesize{Each episode is stored in a directory with the rmb extension.
      Low dimensional data other than images are aggregated and stored in a single HDF5 file.
      RGB images are stored using MP4 compression.
      Depth images are first converted into an RGB representation using a custom encoding scheme and then stored using MP4 compression.}}
  \label{fig:data}
\end{figure}

The collected datasets are stored in a custom data format designed to balance compression efficiency and input output performance, as shown in \figref{fig:data}.
The dataset consists of multiple file types including Hierarchical Data Format 5 (HDF5) and MP4 video files.
Within the code, the data are accessed through a data class interface using data key indices, without exposing the underlying file formats.
Compared with the data formats provided by other frameworks, this format offers several advantages, as described below.

First, in addition to low dimensional data such as robot joint angles, the format can efficiently store RGB images and depth images.
For depth images in particular, the following conversion procedure is introduced to balance precision and data size\footnote{Some parts of this processing utilize the following open source software. \\\scriptsize{\url{https://github.com/vguzov/videoio}}}.
\begin{enumerate}
\item Each pixel depth value is expressed in millimeters and stored as a uint16 value (an unsigned 2 byte integer).
      This representation allows distances up to approximately 65 m to be expressed with 1 mm resolution.
\item The uint16 depth value is split into the upper 8 bits and lower 8 bits, which are assigned to two channels of a uint8 three channel image.
\item This three channel image is treated as an RGB image, and the time series of images is compressed and stored in MP4 video format.
      By converting the data into MP4 video format, standard video compression codecs can be used directly, achieving both high compression ratios and fast input output performance.
\end{enumerate}

Using this method, for example, a depth image sequence with a resolution of 640 $\times$ 480, captured at 22 Hz for 10 seconds, can be stored in approximately 14 MB, while keeping the reconstruction error on the order of several millimeters.
When the data are used, the depth images can be reconstructed by applying the inverse procedure described above.
They can also be converted into three dimensional point clouds using the camera parameters.

Second, within the dataset, the robot state and action are not restricted to a single representation.
Instead, multiple representations are stored in parallel, including both measured values and commanded values.
Specifically, let the joint angles be denoted by $\bm{q}$, the end effector pose by $\bm{p}$, and the forward kinematics function by $\bm{f}_{\mathrm{FK}}$.
Superscripts $\mathrm{msr}$ and $\mathrm{cmd}$ represent measured values and commanded values, respectively.
Then, the following quantities are automatically stored in the dataset:
\[
\bm{p}^{\mathrm{msr}} = \bm{f}_{\mathrm{FK}}(\bm{q}^{\mathrm{msr}}), \quad
\bm{p}^{\mathrm{cmd}} = \bm{f}_{\mathrm{FK}}(\bm{q}^{\mathrm{cmd}})
\]
that is, $\bm{p}^{\mathrm{msr}}, \bm{q}^{\mathrm{msr}}, \bm{p}^{\mathrm{cmd}}, \bm{q}^{\mathrm{cmd}}$.
Furthermore, relative representations defined as differences from the previous time step, namely $\Delta \bm{p}^{\mathrm{msr}}, \Delta \bm{q}^{\mathrm{msr}}, \Delta \bm{p}^{\mathrm{cmd}}, \Delta \bm{q}^{\mathrm{cmd}}$, are also automatically stored.
This design allows flexible switching of the state space $\mathcal{S}$ and action space $\mathcal{A}$ during policy training, enabling systematic comparison and evaluation.

\subsection{Policy} \label{sec:policy}

RoboManipBaselines provides a unified interface for training and rolling out a wide range of policy models using the collected datasets.
In addition to widely used baseline policies such as ACT and Diffusion Policy, the framework supports recent policies based on flow matching, policies that accept 3D point clouds as input, and policies that can flexibly perform multiple tasks based on language instructions.

\begin{description}
\item[\textbf{MLP Policy}]
\mbox{}\\
A simple policy based on a multilayer perceptron (MLP) network.

\item[\textbf{Action Chunking with Transformers (ACT)}~\cite{ALOHA:Zhao:RSS2023}]
\mbox{}\\
A policy that outputs multiple future actions as chunks using a Transformer-based conditional variational autoencoder.

\item[\textbf{Diffusion Policy}~\cite{DiffusionPolicy:Chi:IJRR2024}]
\mbox{}\\
A policy that generates actions using a diffusion model with either a convolutional neural network or a Transformer backbone.

\item[\textbf{Spatial Attention RNN (SARNN)}~\cite{DoorOpen:Ito:SR2022,SARNN:Ichiwara:ICRA2022}]
\mbox{}\\
A policy based on a recurrent neural network that incorporates spatial attention representations and image reconstruction.

\item[\textbf{3D Diffusion Policy}~\cite{3DDiffusionPolicy:Ze:RSS2024}]
\mbox{}\\
A policy that takes 3D point clouds as input and generates actions using a diffusion model.

\item[\textbf{ManiFlow Policy}~\cite{ManiFlowPolicy:Yan:CoRL2025}, \textbf{Flow Policy}~\cite{FlowPolicy:Zhang:AAAI2025}]
\mbox{}\\
Policies that generate actions based on flow matching using camera images or 3D point clouds as input.

\item[\textbf{Multi-Task ACT (MT-ACT)}~\cite{MtAct:Bharadhwaj:ICRA2024}]
\mbox{}\\
An extension of ACT that accepts language instructions as input and supports multiple tasks.

\item[\textbf{$\bm{\pi_0}$}~\cite{pi0:Black:arXiv2024}]
\mbox{}\\
A policy that generates actions based on flow matching using PaliGemma~\cite{PaliGemma:Beyer:arXiv2024} as the backbone vision-language model (VLM).

\item[\textbf{GR00T}~\cite{GR00T:Bjorck:arXiv2025}]
\mbox{}\\
A policy that generates actions using a diffusion Transformer with Eagle-2~\cite{Eagle2:Li:arXiv2025} as the backbone VLM.
\end{description}

To preserve the original implementations of prior work, the policy model definitions directly reuse the code released by the respective authors.
In contrast, dataset loading, management of trained parameters, and input output processing for rollout are implemented in a unified and reusable manner across all policies.
Thanks to the environment abstraction, policy deployment is unified across both real and simulated robots.

When adding a new policy, users only need to inherit the abstract classes for training and rollout and override a minimal set of methods related to the policy model.
As with the addition of new environments, RoboManipBaselines can be extended by inheriting its open source code from an external repository.
This design allows custom policies to be implemented independently without modifying the original framework.

To accelerate data access during policy training, the framework provides an option to load the entire dataset into CPU memory.
After the dataset has been accessed once, subsequent accesses no longer require file operations, enabling significantly faster data retrieval.
On machines with sufficient CPU memory, enabling this feature can accelerate policy training by several times.

\section{Basic Usage}

This section presents the basic usage of RoboManipBaselines.
Detailed and comprehensive documentation is available in the GitHub repository\footnote{\scriptsize{\url{https://github.com/isri-aist/RoboManipBaselines/blob/master/README.md}}}.
Here, we illustrate the flexibility of the framework by outlining the commands required to execute a typical imitation learning workflow.
The following example corresponds to a standard workflow consisting of data collection, policy training, and policy rollout.
The framework is implemented in Python and can be installed via the {\small \texttt{pip}} command in a Linux environment.
In practice, the execution requires simulator dependencies (e.g., MuJoCo) and an appropriate runtime environment.
Detailed setup instructions, including dependency installation and troubleshooting tips, are provided in the project repository.


\subsection{Data Collection}

To collect demonstration data for a cable manipulation task by teleoperating a Universal Robots UR5e in the MuJoCo simulator using keyboard input, the following command can be executed.
\begin{codeblock}
\begin{verbatim}
$ python ./bin/Teleop.py MujocoUR5eCable \
      --input_device keyboard
\end{verbatim}
\end{codeblock}
By changing the argument \texttt{MujocoUR5eCable} to \texttt{RealXarm7Demo}, data collection can be performed on a real UFACTORY xArm~7.
Similarly, changing the argument \texttt{keyboard} to \texttt{spacemouse} enables teleoperation using a 3D mouse.

\subsection{Policy Training}

If the collected dataset is stored in \texttt{<dataset\_dir>}, the following command can be executed to train an ACT policy using this dataset.
\begin{codeblock}
\begin{verbatim}
$ python ./bin/Train.py Act \
    --dataset_dir <dataset_dir>
\end{verbatim}
\end{codeblock}
Replacing the argument \texttt{Act} with \texttt{Mlp} or \texttt{DiffusionPolicy} trains the corresponding policy models.
The observation and action representations can be switched using the \texttt{--state\_keys} and \texttt{--action\_keys} options, for example between joint angles and end effector poses or between absolute positions and relative displacements.
When the dataset contains images from multiple cameras, the cameras used for training can be specified using the \texttt{--camera\_names} option.

\subsection{Policy Rollout}

To rollout the trained policy in the same environment used for data collection, the following command can be executed.
\begin{codeblock}
\begin{verbatim}
$ python ./bin/Rollout.py Act MujocoUR5eCable \
      --checkpoint <checkpoint_path>
\end{verbatim}
\end{codeblock}
\texttt{<checkpoint\_path>} specifies the file containing the trained parameters of the ACT policy.
The option settings used during training are stored in this file, so they do not need to be specified again at rollout time.

\section{Benchmark Evaluation}

\begin{figure*}[tb]
  \centering
  \includegraphics[width=0.11\textwidth]{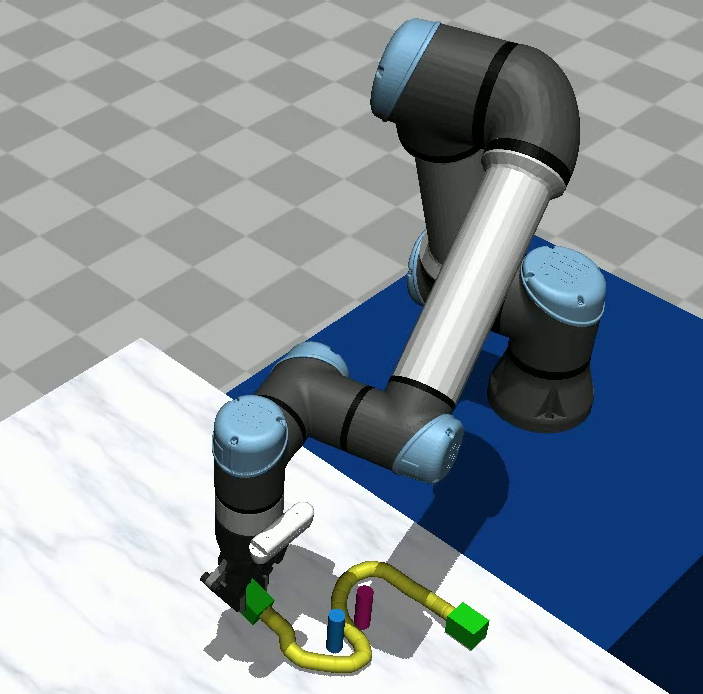}
  \includegraphics[width=0.11\textwidth]{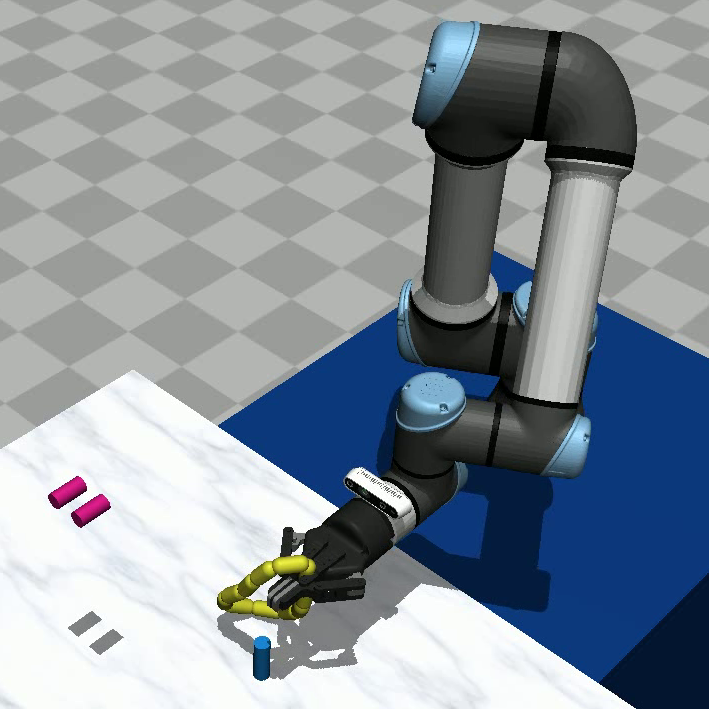}
  \includegraphics[width=0.11\textwidth]{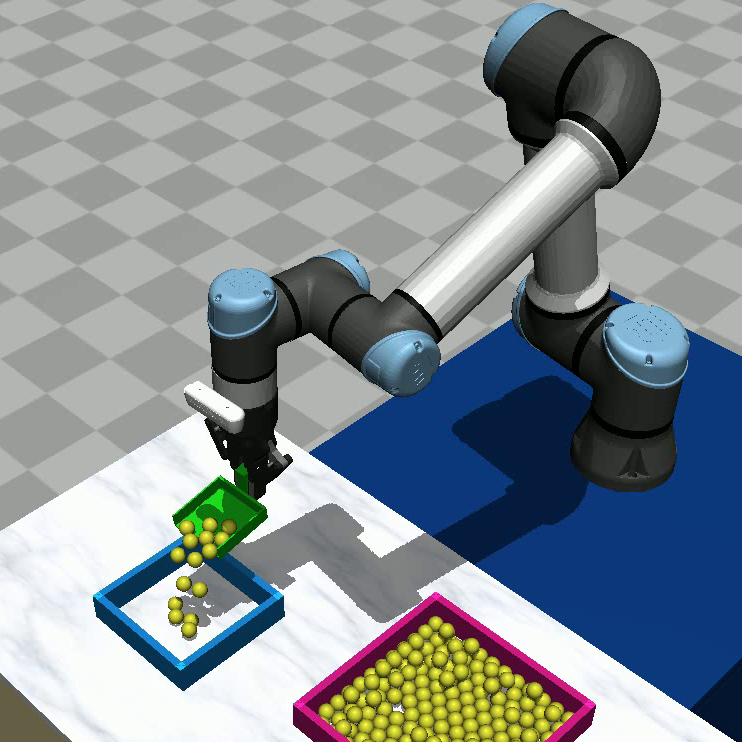}
  \includegraphics[width=0.11\textwidth]{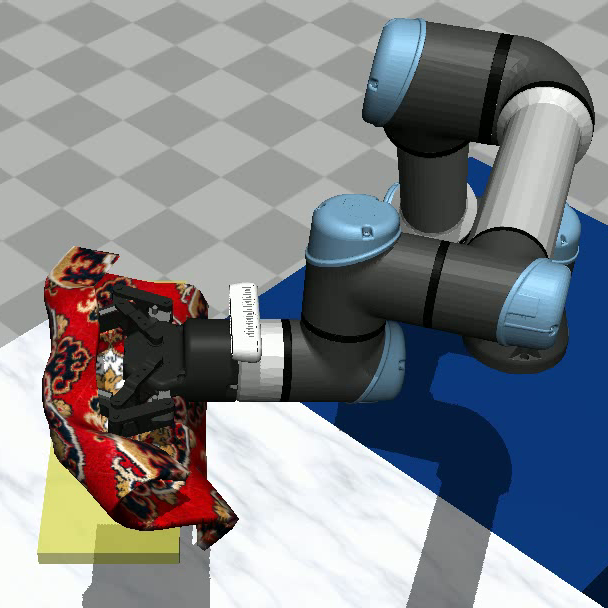}
  \includegraphics[width=0.11\textwidth]{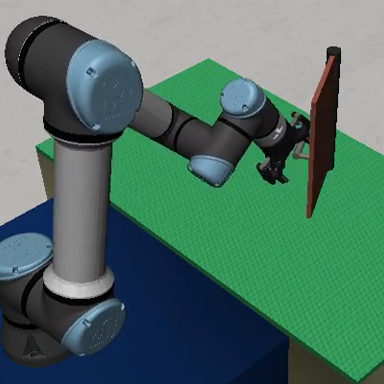}
  \includegraphics[width=0.11\textwidth]{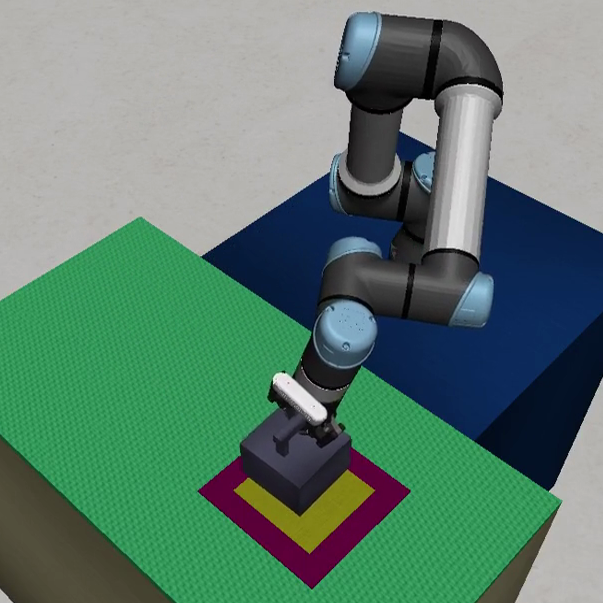}
  \includegraphics[width=0.11\textwidth]{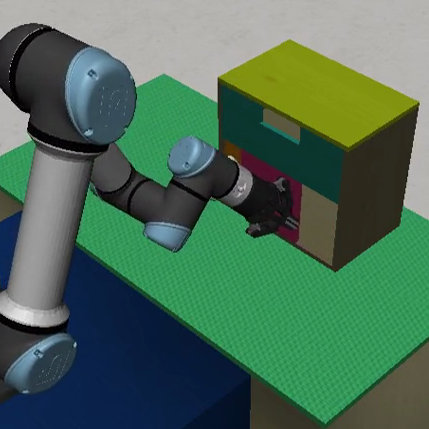}
  \includegraphics[width=0.11\textwidth]{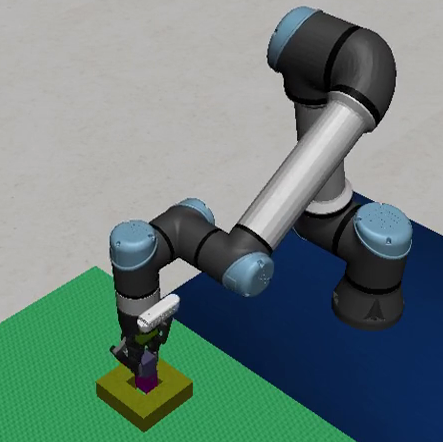}\\
  \begin{minipage}{0.11\textwidth}
    \begin{center} \footnotesize Cable \end{center}
  \end{minipage}
  \begin{minipage}{0.11\textwidth}
    \begin{center} \footnotesize Ring \end{center}
  \end{minipage}
  \begin{minipage}{0.11\textwidth}
    \begin{center} \footnotesize Particle \end{center}
  \end{minipage}
  \begin{minipage}{0.11\textwidth}
    \begin{center} \footnotesize Cloth \end{center}
  \end{minipage}
  \begin{minipage}{0.11\textwidth}
    \begin{center} \footnotesize Door \end{center}
  \end{minipage}
  \begin{minipage}{0.11\textwidth}
    \begin{center} \footnotesize Toolbox \end{center}
  \end{minipage}
  \begin{minipage}{0.11\textwidth}
    \begin{center} \footnotesize Cabinet \end{center}
  \end{minipage}
  \begin{minipage}{0.11\textwidth}
    \begin{center} \footnotesize Insert \end{center}
  \end{minipage}\\
  \begin{minipage}{1.0\textwidth}
    \begin{center} \footnotesize (A) Simulation environments \end{center}
  \end{minipage}\\
  \vspace{3mm}
  \includegraphics[width=0.11\textwidth]{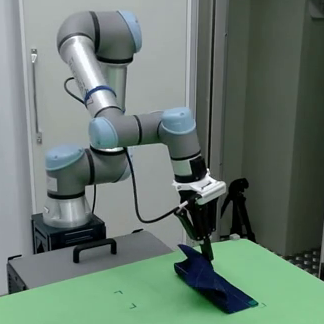}
  \includegraphics[width=0.11\textwidth]{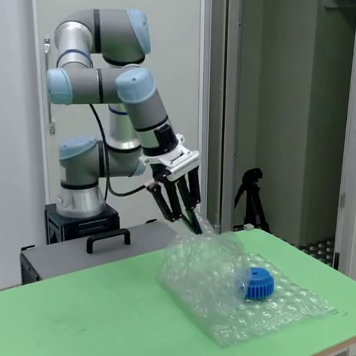}
  \includegraphics[width=0.11\textwidth]{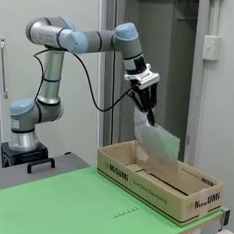}
  \includegraphics[width=0.11\textwidth]{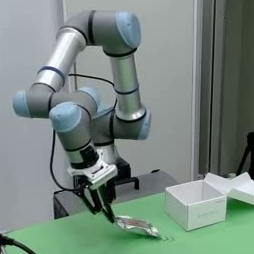}
  \includegraphics[width=0.11\textwidth]{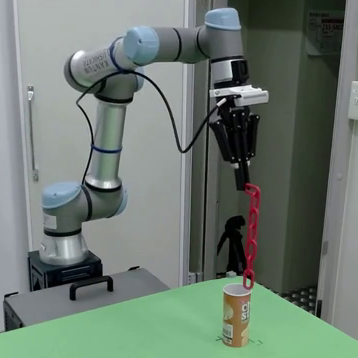}
  \includegraphics[width=0.11\textwidth]{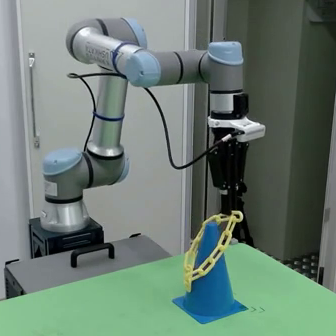}
  \includegraphics[width=0.11\textwidth]{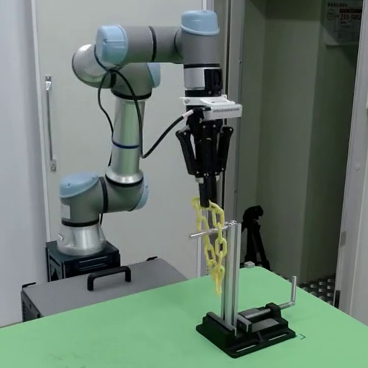}
  \includegraphics[width=0.11\textwidth]{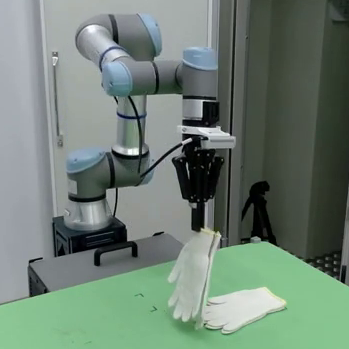}\\
  \begin{minipage}{0.11\textwidth}
    \begin{center} \footnotesize Handkerchief \end{center}
  \end{minipage}
  \begin{minipage}{0.11\textwidth}
    \begin{center} \footnotesize Sheet \end{center}
  \end{minipage}
  \begin{minipage}{0.11\textwidth}
    \begin{center} \footnotesize Bag(L) \end{center}
  \end{minipage}
  \begin{minipage}{0.11\textwidth}
    \begin{center} \footnotesize Bag(S) \end{center}
  \end{minipage}
  \begin{minipage}{0.11\textwidth}
    \begin{center} \footnotesize ChainInTube \end{center}
  \end{minipage}
  \begin{minipage}{0.11\textwidth}
    \begin{center} \footnotesize RingOnCone \end{center}
  \end{minipage}
  \begin{minipage}{0.11\textwidth}
    \begin{center} \footnotesize RingOnHook \end{center}
  \end{minipage}
  \begin{minipage}{0.11\textwidth}
    \begin{center} \footnotesize Gloves \end{center}
  \end{minipage}\\
  \begin{minipage}{1.0\textwidth}
    \begin{center} \footnotesize (B) Real-world environments \end{center}
  \end{minipage}
  \caption{Manipulation tasks used in the benchmark evaluation.}
  \label{fig:eval-envs}
\end{figure*}

\begin{table*}[tb]
\centering
\renewcommand{\arraystretch}{1.2}
\caption{Task success rates in simulation environments.}
\vspace{-2mm}
\label{tab:eval-sim-all}
\begin{tabular}{l|cccccccc|c}
\toprule
Policy $\backslash$ Task & Cable & Ring & Particle & Cloth & Door & Toolbox & Cabinet & Insert & Average \\
\midrule
MLP Policy & 0\% {\tiny ($\pm$0)} & 7\% {\tiny ($\pm$15)} & 0\% {\tiny ($\pm$0)} & 27\% {\tiny ($\pm$19)} & 10\% {\tiny ($\pm$9)} & 47\% {\tiny ($\pm$38)} & 23\% {\tiny ($\pm$22)} & 17\% {\tiny ($\pm$20)} & 16.4\% {\tiny ($\pm$15)} \\
ACT & 73\% {\tiny ($\pm$32)} & \textbf{45\% {\tiny ($\pm$30)}} & 27\% {\tiny ($\pm$38)} & 53\% {\tiny ($\pm$34)} & \textbf{55\% {\tiny ($\pm$25)}} & 57\% {\tiny ($\pm$15)} & 13\% {\tiny ($\pm$18)} & \textbf{37\% {\tiny ($\pm$18)}} & 45.0\% {\tiny ($\pm$18)} \\
Diffusion Policy & 42\% {\tiny ($\pm$8)} & 0\% {\tiny ($\pm$0)} & 42\% {\tiny ($\pm$11)} & \textbf{97\% {\tiny ($\pm$7)}} & 38\% {\tiny ($\pm$20)} & \textbf{77\% {\tiny ($\pm$9)}} & \textbf{87\% {\tiny ($\pm$14)}} & 33\% {\tiny ($\pm$12)} & \textbf{52.0\% {\tiny ($\pm$30)}} \\
SARNN & \textbf{77\% {\tiny ($\pm$21)}} & 35\% {\tiny ($\pm$25)} & \textbf{63\% {\tiny ($\pm$24)}} & 85\% {\tiny ($\pm$12)} & 2\% {\tiny ($\pm$5)} & 47\% {\tiny ($\pm$7)} & 60\% {\tiny ($\pm$25)} & \textbf{37\% {\tiny ($\pm$7)}} & 50.8\% {\tiny ($\pm$25)} \\
\bottomrule
\end{tabular}\\
\vspace{2mm}
\begin{minipage}{1.8\columnwidth}
\footnotesize{
  Four policies are evaluated on eight manipulation tasks in simulation.
  Each value represents the mean success rate ($\pm$ standard deviation) over five random seeds.
}
\end{minipage}
\end{table*}

\begin{table*}[tb]
\centering
\renewcommand{\arraystretch}{1.2}
\caption{Task success rates of ACT under different camera configurations.}
\vspace{-2mm}
\label{tab:eval-sim-camera}
\begin{tabular}{l|cccccccc|c}
\toprule
Cameras $\backslash$ Task & Cable & Ring & Particle & Cloth & Door & Toolbox & Cabinet & Insert & Average \\
\midrule
Front & \textbf{73\% {\tiny ($\pm$32)}} & 45\% {\tiny ($\pm$30)} & 27\% {\tiny ($\pm$38)} & 53\% {\tiny ($\pm$34)} & \textbf{55\% {\tiny ($\pm$25)}} & 57\% {\tiny ($\pm$15)} & 13\% {\tiny ($\pm$18)} & 37\% {\tiny ($\pm$18)} & 45.0\% {\tiny ($\pm$18)} \\
Side & \textbf{73\% {\tiny ($\pm$15)}} & 57\% {\tiny ($\pm$40)} & 43\% {\tiny ($\pm$38)} & 67\% {\tiny ($\pm$39)} & 7\% {\tiny ($\pm$15)} & 63\% {\tiny ($\pm$36)} & \textbf{20\% {\tiny ($\pm$30)}} & 50\% {\tiny ($\pm$12)} & 47.5\% {\tiny ($\pm$22)} \\
Hand & 0\% {\tiny ($\pm$0)} & 23\% {\tiny ($\pm$9)} & 0\% {\tiny ($\pm$0)} & 17\% {\tiny ($\pm$29)} & 17\% {\tiny ($\pm$24)} & 40\% {\tiny ($\pm$37)} & 0\% {\tiny ($\pm$0)} & 40\% {\tiny ($\pm$9)} & 17.1\% {\tiny ($\pm$26)} \\
Front+Side & 47\% {\tiny ($\pm$32)} & \textbf{73\% {\tiny ($\pm$19)}} & \textbf{60\% {\tiny ($\pm$48)}} & \textbf{70\% {\tiny ($\pm$32)}} & 27\% {\tiny ($\pm$35)} & 80\% {\tiny ($\pm$14)} & 13\% {\tiny ($\pm$22)} & 47\% {\tiny ($\pm$7)} & \textbf{52.1\% {\tiny ($\pm$22)}} \\
Front+Hand & 10\% {\tiny ($\pm$22)} & 63\% {\tiny ($\pm$30)} & 0\% {\tiny ($\pm$0)} & 10\% {\tiny ($\pm$15)} & 23\% {\tiny ($\pm$25)} & 77\% {\tiny ($\pm$19)} & 0\% {\tiny ($\pm$0)} & 30\% {\tiny ($\pm$27)} & 26.6\% {\tiny ($\pm$33)} \\
Side+Hand & 17\% {\tiny ($\pm$17)} & \textbf{73\% {\tiny ($\pm$25)}} & 3\% {\tiny ($\pm$7)} & 60\% {\tiny ($\pm$42)} & 3\% {\tiny ($\pm$7)} & 73\% {\tiny ($\pm$19)} & 0\% {\tiny ($\pm$0)} & 47\% {\tiny ($\pm$30)} & 34.5\% {\tiny ($\pm$33)} \\
Front+Side+Hand & 40\% {\tiny ($\pm$38)} & 57\% {\tiny ($\pm$15)} & 3\% {\tiny ($\pm$7)} & 30\% {\tiny ($\pm$27)} & 20\% {\tiny ($\pm$22)} & \textbf{90\% {\tiny ($\pm$15)}} & 17\% {\tiny ($\pm$29)} & \textbf{63\% {\tiny ($\pm$7)}} & 40.0\% {\tiny ($\pm$31)} \\
\bottomrule
\end{tabular}\\
\vspace{2mm}
\begin{minipage}{1.8\columnwidth}
  \footnotesize{
    ACT is evaluated on eight manipulation tasks in simulation environments under different camera configurations.
    Seven configurations are compared using different combinations of the front, side, and wrist-mounted cameras (Front, Side, and Hand).
    Each value represents the mean success rate ($\pm$ standard deviation) over five random seeds.
  }
\end{minipage}
\end{table*}

\begin{table*}[tb]
\centering
\renewcommand{\arraystretch}{1.2}
\caption{Task success rates in real-world environments.}
\vspace{-2mm}
\label{tab:eval-real}
\begin{tabular}{l|cccccccc|c}
\toprule
Policy $\backslash$ Task & Handkerchief & Sheet & Bag(L) & Bag(S) & ChainInTube & RingOnCone & RingOnHook & Gloves & Average \\
\midrule
ACT & 67\% & 33\% & \textbf{100\%} & \textbf{100\%} & 17\% & 0\% & 33\% & 0\% & 43.8\% {\tiny ($\pm$38)} \\
Diffusion Policy & \textbf{100\%} & \textbf{67\%} & 17\% & 50\% & 17\% & \textbf{67\%} & \textbf{67\%} & 0\% & \textbf{47.9\% {\tiny ($\pm$32)}} \\
SARNN & 83\% & \textbf{67\%} & 0\% & 0\% & \textbf{100\%} & 0\% & 50\% & \textbf{17\%} & 39.6\% {\tiny ($\pm$38)} \\
\bottomrule
\end{tabular}
\end{table*}

\begin{figure*}[tb]
  \centering
  \includegraphics[width=1.0\textwidth]{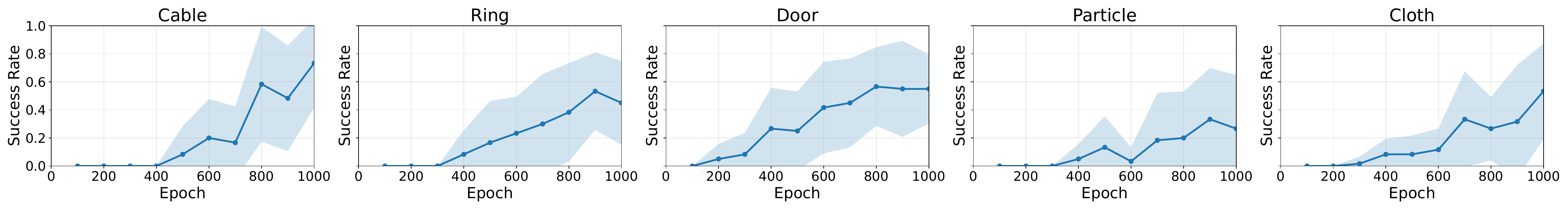}\\
  \vspace{-1mm}
  \begin{minipage}{1.0\textwidth}
    \begin{center} \footnotesize (A) ACT \end{center}
  \end{minipage}\\
  \vspace{3mm}
  \includegraphics[width=1.0\textwidth]{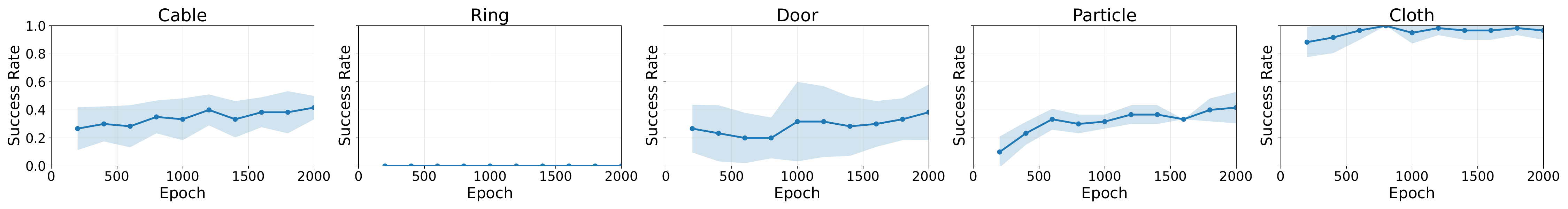}\\
  \vspace{-1mm}
  \begin{minipage}{1.0\textwidth}
    \begin{center} \footnotesize (B) Diffusion Policy \end{center}
  \end{minipage}\\
  \vspace{3mm}
  \includegraphics[width=1.0\textwidth]{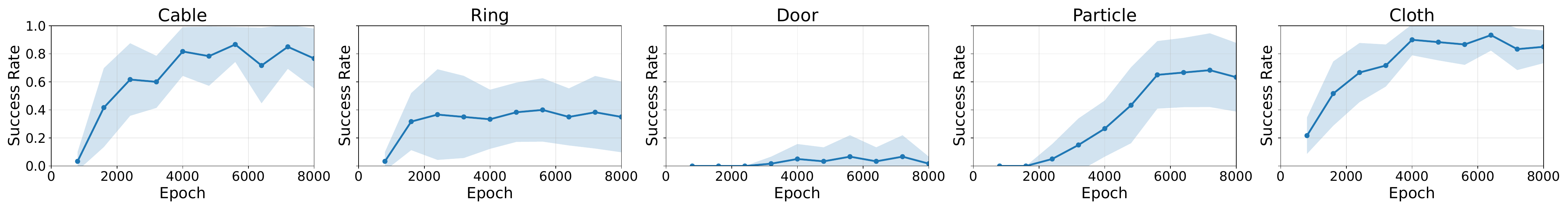}\\
  \vspace{-1mm}
  \begin{minipage}{1.0\textwidth}
    \begin{center} \footnotesize (C) SARNN \end{center}
  \end{minipage}\\
  \caption{
    Success rates over training epochs.\newline
    \footnotesize{Three policies are evaluated on five manipulation tasks in simulation environments.
      Evaluation is performed over five random seeds, and the shaded regions indicate the standard deviation.}
  }
  \label{fig:eval-sim-epoch}
\end{figure*}

\begin{figure}[tb]
  \centering
  \includegraphics[width=1.0\columnwidth]{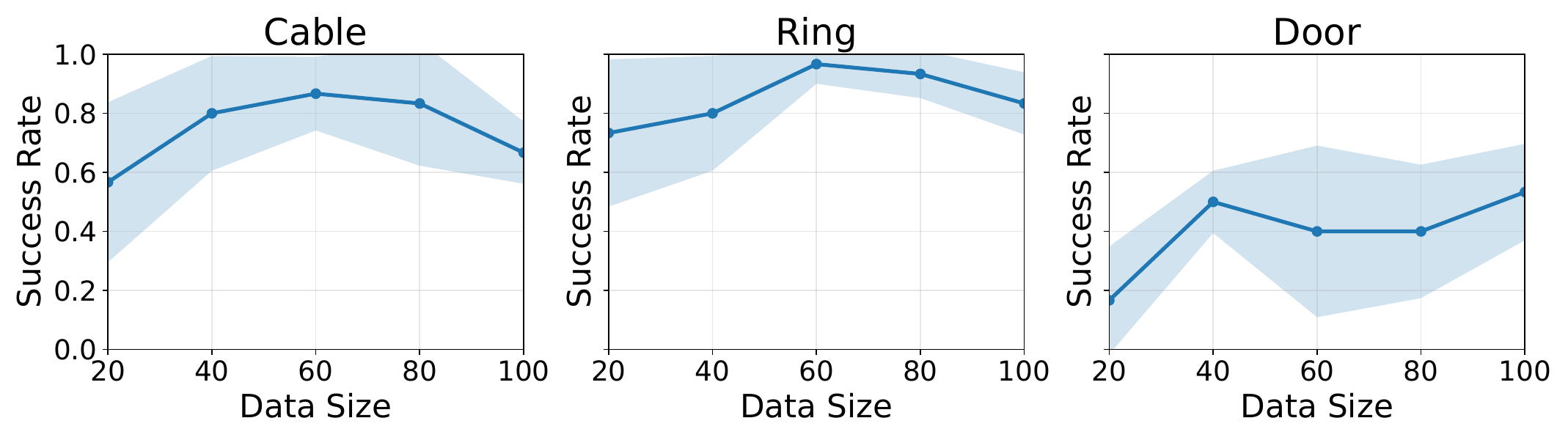}\\
  \vspace{-1mm}
  \begin{minipage}{1.0\columnwidth}
    \begin{center} \footnotesize (A) ACT \end{center}
  \end{minipage}\\
  \vspace{3mm}
  \includegraphics[width=1.0\columnwidth]{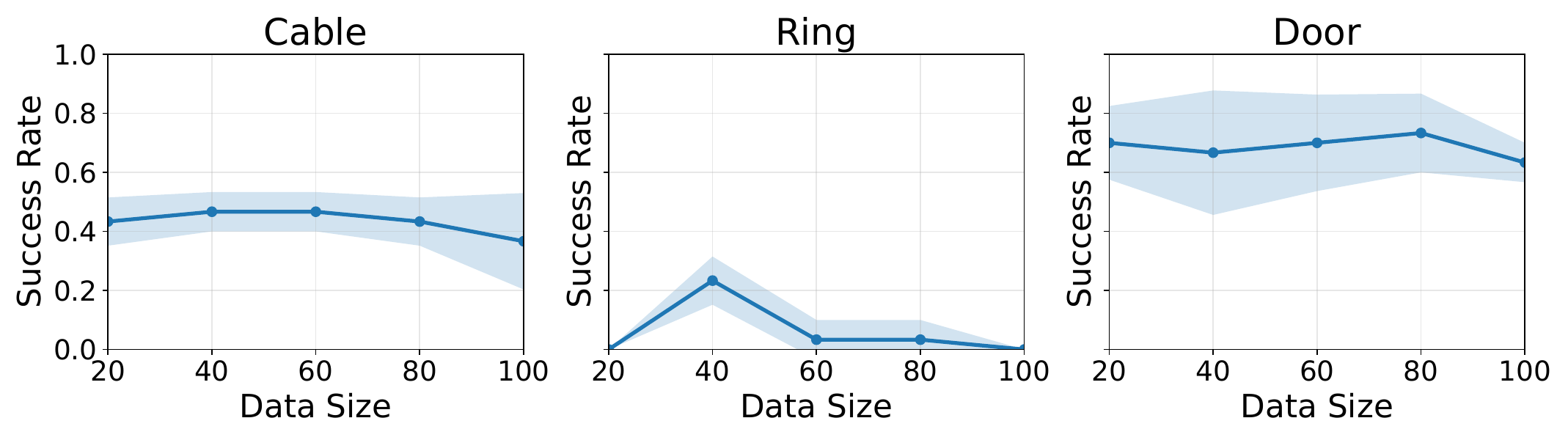}\\
  \vspace{-1mm}
  \begin{minipage}{1.0\columnwidth}
    \begin{center} \footnotesize (B) Diffusion Policy \end{center}
  \end{minipage}\\
  \vspace{3mm}
  \includegraphics[width=1.0\columnwidth]{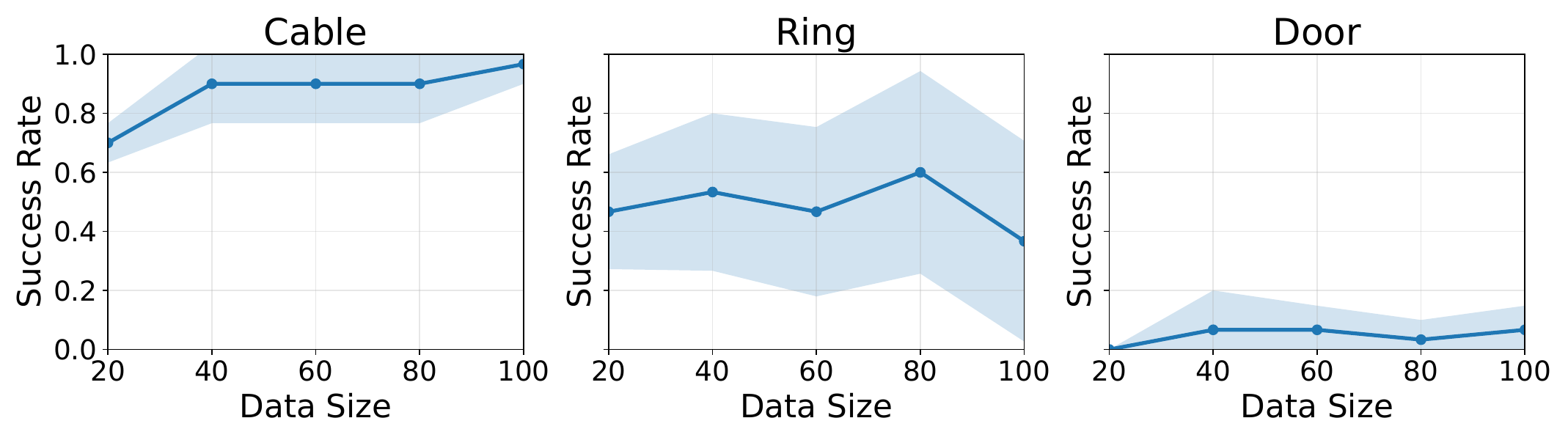}\\
  \vspace{-1mm}
  \begin{minipage}{1.0\columnwidth}
    \begin{center} \footnotesize (C) SARNN \end{center}
  \end{minipage}\\
  \caption{
    Success rates over the number of training samples.\newline
    \footnotesize{Three policies are evaluated on three manipulation tasks in simulation environments.
      Evaluation is performed over five random seeds, and the shaded regions indicate the standard deviation.}
  }
  \label{fig:eval-sim-datasize}
\end{figure}

\subsection{Evaluation in Simulation}

\subsubsection{Task Setup and Experimental Results}

Using RoboManipBaselines, we evaluated various manipulation tasks performed by a Universal Robots UR5e in the MuJoCo environments.
\figref{fig:eval-envs}~(A) shows the eight manipulation tasks used for evaluation.

\textbf{Cable}: Route a flexible cable between two poles and arrange it in an S shape.

\textbf{Ring}: Pick up a ring hanging on a hook and place it onto a pole.

\textbf{Particle}: Use a scoop to transfer particles from one container to another.

\textbf{Cloth}: Lift and roll up a hanging cloth.

\textbf{Door}: Grasp the doorknob and open the door.

\textbf{Toolbox}: Grasp the handle of a toolbox and move it onto a mat.

\textbf{Cabinet}: Hook the gripper onto the handle and slide the cabinet door open.

\textbf{Insert}: Insert a square peg into a hole.

Cable, Ring, Particle, and Cloth involve deformable objects, while Door and Cabinet involve articulated objects.
All tasks require advanced manipulation skills.

\tabref{tab:eval-sim-all} presents the evaluation results for four policies: MLP Policy, ACT, Diffusion Policy, and SARNN.
A brief description of each policy is given in Section~\ref{sec:policy}.
The policy achieving the highest success rate varies across tasks among ACT, Diffusion Policy, and SARNN.
Compared with the simple MLP Policy, all three policies achieve substantially higher success rates.
This suggests that architectural components such as action chunking and diffusion-based action generation play an important role in task success.

In terms of the average performance across all tasks, Diffusion Policy achieves the highest score by a small margin.
SARNN, on the other hand, uses a relatively shallow network architecture and does not require a GPU during inference.
Despite this simplicity, the combination of spatial attention representations, image reconstruction, and a recurrent neural network enables SARNN to achieve performance comparable to computationally intensive policies such as ACT and Diffusion Policy.













\subsubsection{Comparison of Camera Number and Viewpoints}

\tabref{tab:eval-sim-camera} shows the evaluation results of ACT on the same eight simulation tasks as in \tabref{tab:eval-sim-all} when varying the number and viewpoints of cameras.
Seven camera configurations are compared using different combinations of the front view of the workspace (Front), the side view (Side), and the wrist-mounted view of the robot (Hand).

The results show that the highest success rate is obtained when using two cameras, Front and Side, while the lowest success rate occurs when using only the Hand camera.
Configurations including the Hand camera show relatively lower success rates, even when combined with the Front or Side cameras.
In these tasks, images from the Hand camera often exhibit large viewpoint changes due to robot motion, which may make policy learning more difficult.



\subsubsection{Learning Curves and Data Efficiency}

To analyze the training performance of imitation learning policies, we evaluated success rates with respect to training epochs and the number of training episodes.
Three policies were considered in this analysis: ACT, Diffusion Policy, and SARNN.

First, the success rates as a function of training epochs are shown in \figref{fig:eval-sim-epoch}.
Success rates generally increase as training progresses, while the convergence speed and final performance vary depending on the policy and task.

Next, the success rates as a function of the number of training episodes are shown in \figref{fig:eval-sim-datasize}.
For many policy and task combinations, task success rates do not increase monotonically as the number of training episodes grows, and in some cases even decrease.
These results suggest that, in single-task settings, these policies can achieve sufficient performance with relatively small numbers of demonstrations.
They also indicate that the increased trajectory variability introduced by larger datasets does not necessarily improve learning.




\subsection{Evaluation in Real-world Environments}

Using RoboManipBaselines, we evaluated various manipulation tasks performed by a Universal Robots UR5e in a real-world setup.
\figref{fig:eval-envs}~(B) shows the eight manipulation tasks used for evaluation.

\textbf{Handkerchief}: Fold a square handkerchief in half.

\textbf{Sheet}: Fold a transparent cushioning sheet in half.

\textbf{Bag(L)}: Place a large aluminum foil bag in a box.

\textbf{Bag(S)}: Place a small aluminum foil bag in a box.

\textbf{ChainInTube}: Insert a plastic chain into a tube.

\textbf{RingOnCone}: Place a plastic chain ring on a cone.

\textbf{RingOnHook}: Place a plastic chain ring on a metal hook.

\textbf{Gloves}: Place one glove on top of the other.

These tasks involve deformable objects as well as transparent and reflective objects, and all require advanced manipulation skills.

\tabref{tab:eval-real} presents the evaluation results for three policies: ACT, Diffusion Policy, and SARNN.
A brief description of each policy is given in Section~\ref{sec:policy}.
The policy achieving the highest success rate varies across tasks.
In terms of the average performance across all tasks, Diffusion Policy achieves the highest score by a small margin, consistent with the simulation results.

These results demonstrate that RoboManipBaselines enables unified benchmarking from simulation environments to real-world robotic systems.













\section{Research Applications}

This section presents several research applications of RoboManipBaselines.
To illustrate its versatility, we introduce diverse examples including data augmentation, integration with tactile models, interactive robotic systems, evaluation of 3D sensors, and hardware extensions.


\subsection{Data Augmentation}

\begin{figure}[tb]
  \centering
  \includegraphics[width=0.95\columnwidth]{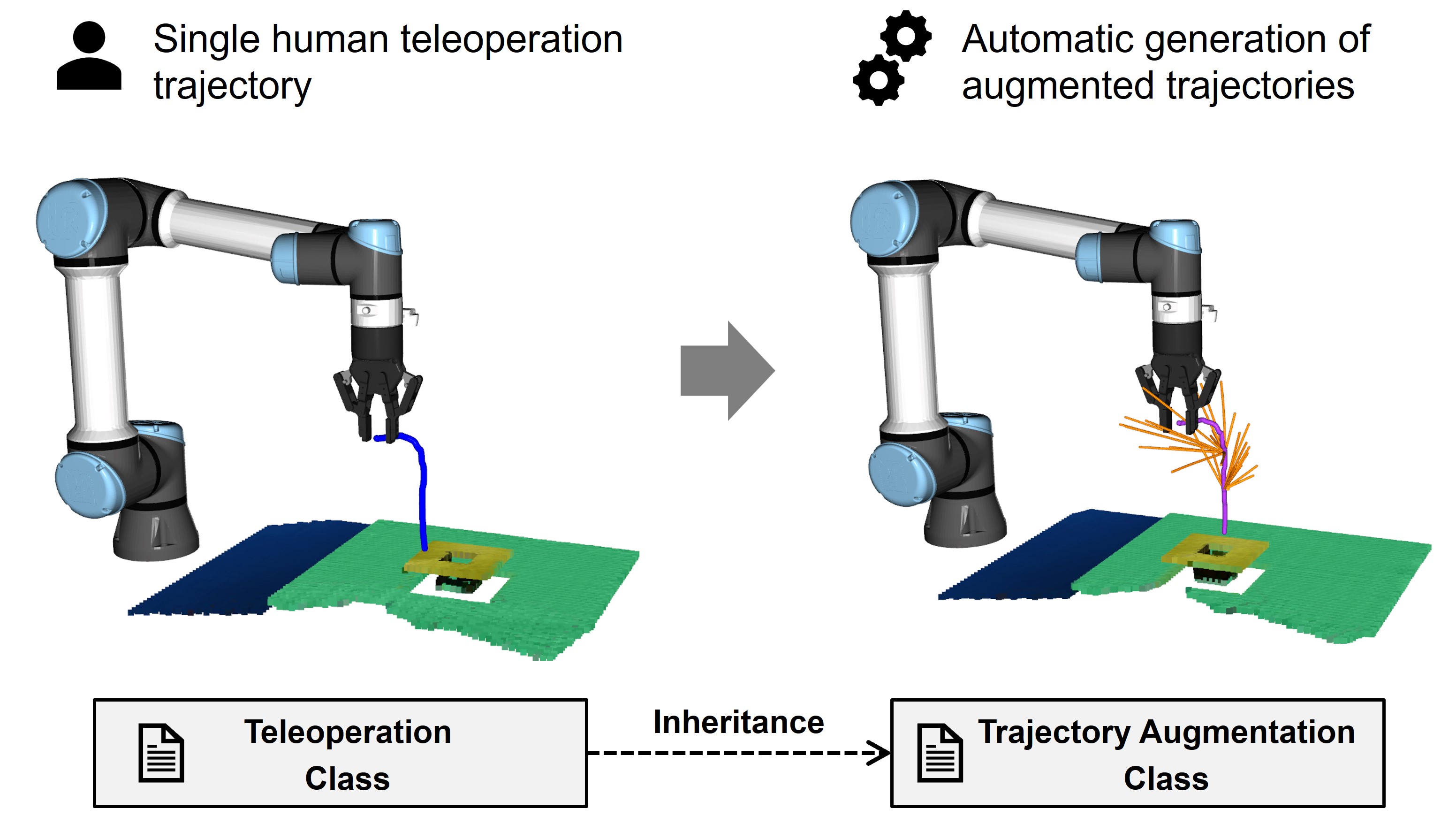}
  \caption{
    Data augmentation.\newline
    \footnotesize{Starting from a single teleoperated demonstration, additional trajectories are automatically generated around the reference trajectory.
      The trajectory augmentation class is implemented by inheriting the teleoperation class in RoboManipBaselines.}
  }
  \label{fig:aug}
\end{figure}

We present a data augmentation study for imitation learning using RoboManipBaselines.
Details of the method and quantitative evaluation results are reported in a separate publication~\cite{SART:Oh:AR2026}.
Here we focus on how RoboManipBaselines was effectively utilized as a general framework.

The data augmentation pipeline used in this study consists of the following three phases.
\begin{itemize}
  \item \textbf{Phase 1:} A human operator performs teleoperation once to collect a reference demonstration trajectory.
  \item \textbf{Phase 2:} The robot autonomously generates additional trajectories around the collected trajectory to augment the dataset.
  \item \textbf{Phase 3:} An imitation learning policy is trained on the augmented dataset and deployed in the real environment.
\end{itemize}
Using this pipeline, the success rate across multiple manipulation tasks in both real-world and simulation environments improved from 36\% to 82\% compared with training without data augmentation.

Phase~1 and Phase~3 were implemented using the standard functionalities of RoboManipBaselines: teleoperation with a 3D mouse for Phase~1, and training and rollout of an MLP policy for Phase~3.
Phase~2 was implemented by inheriting the teleoperation class to generate augmented trajectories, as illustrated in \figref{fig:aug}.
By extending RoboManipBaselines through inheritance, only the minimal functionality required for data augmentation was implemented in a separate project without copying or modifying the original framework\footnote{The code is available at:\\\scriptsize{\url{https://github.com/isri-aist/RoboManipAug}}}.
This approach enables research-specific extensions while preserving the generality of RoboManipBaselines.




\subsection{Integration with a Tactile Foundation Model}

\begin{figure}[tb]
  \centering
  \includegraphics[width=0.85\columnwidth]{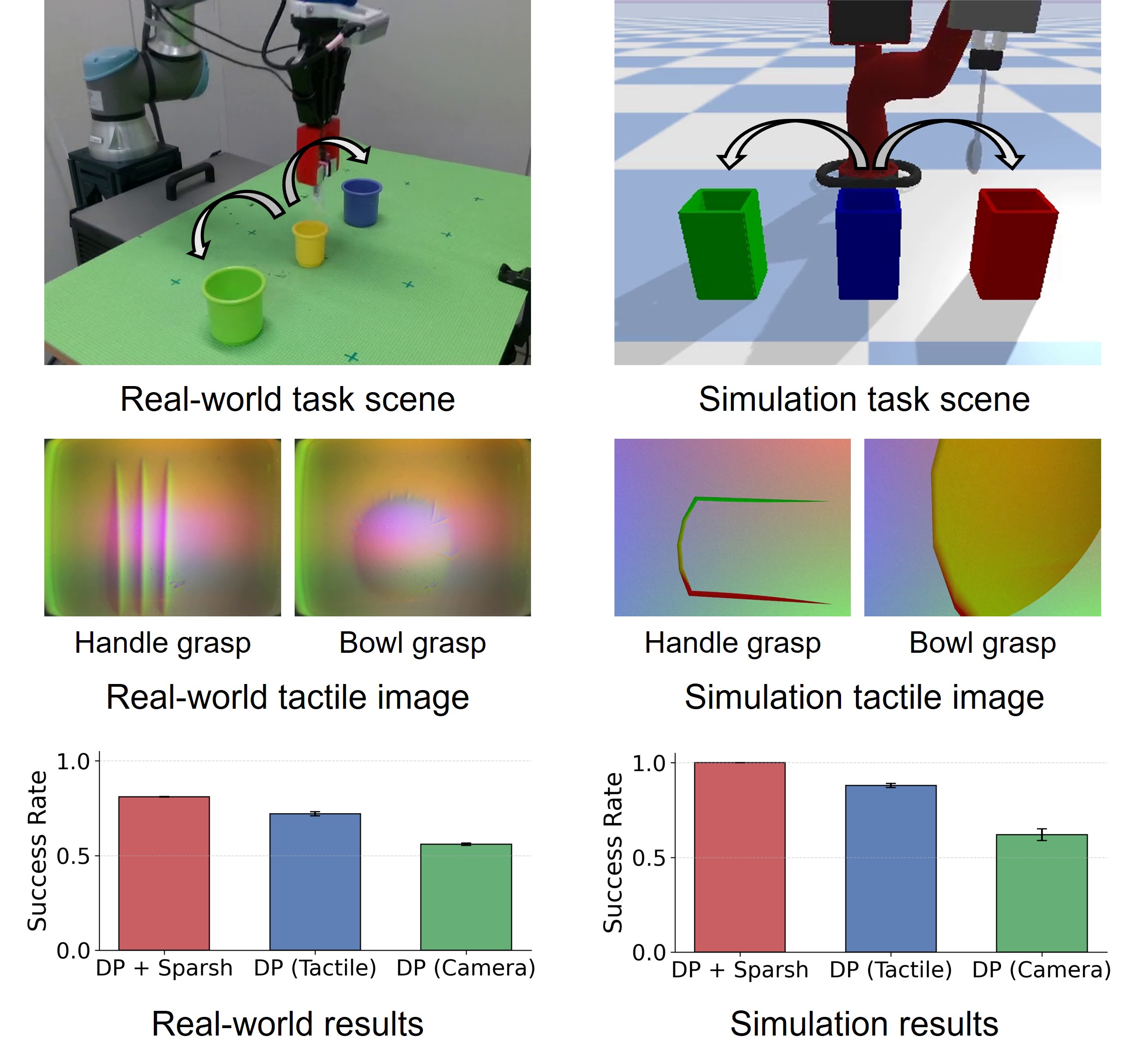}
  \caption{
    Integration with a tactile foundation model.\newline
    \footnotesize{In this task, a spoon is randomly placed in a central cup with either the handle or bowl side facing upward. The robot picks up the spoon and places it into one of two cups depending on which side is grasped. Clear differences can be observed in the tactile images obtained from the gripper-mounted tactile sensor depending on whether the handle or bowl side is grasped. We evaluate the integration of the tactile foundation model Sparsh~\cite{Sparsh:Higuera:CoRL2024} with Diffusion Policy. Compared with models using only camera observations (DP (Camera)) or naively encoding tactile images in the same way as RGB images (DP (Tactile)), the integrated model (DP + Sparsh) achieves higher success rates. By integrating the tactile simulator TACTO~\cite{TACTO:Wang:RAL2022} into RoboManipBaselines, the same experiments can be conducted transparently in both real-world and simulation environments.}
  }
  \label{fig:tactile}
\end{figure}

We present an evaluation of integrating imitation learning with a pretrained tactile foundation model using RoboManipBaselines.
Tactile sensing plays an important role in imitation learning for fine-grained manipulation tasks~\cite{TactileAloha:Gu:RAL2025}, yet it remains unclear how tactile observations should be encoded into latent representations.
We focus on tactile foundation models trained with supervised learning on large-scale tactile datasets, whose effectiveness has been demonstrated in various downstream recognition and generation tasks.
Specifically, we use the latent representation produced by the tactile foundation model Sparsh~\cite{Sparsh:Higuera:CoRL2024} as the conditioning input of Diffusion Policy~\cite{DiffusionPolicy:Chi:IJRR2024}.
Experiments are conducted on a manipulation task requiring tactile perception using a manipulator equipped with a GelSight optical tactile sensor.
As shown in \figref{fig:tactile}, the model integrated with the Sparsh tactile encoder achieves higher success rates than a naive Diffusion Policy model.

This study builds on the Diffusion Policy implementation in RoboManipBaselines (model, training, and rollout), with only minimal additional code for integration with Sparsh implemented as a separate project\footnote{The code will be released after the full evaluation and analysis are completed.}.
By integrating the TACTO simulator~\cite{TACTO:Wang:RAL2022} into RoboManipBaselines, the same model could be evaluated in both real-world and simulation environments using the same interface.



\subsection{Interactive Robotic System}

\begin{figure}[tb]
  \centering
  \includegraphics[width=0.85\columnwidth]{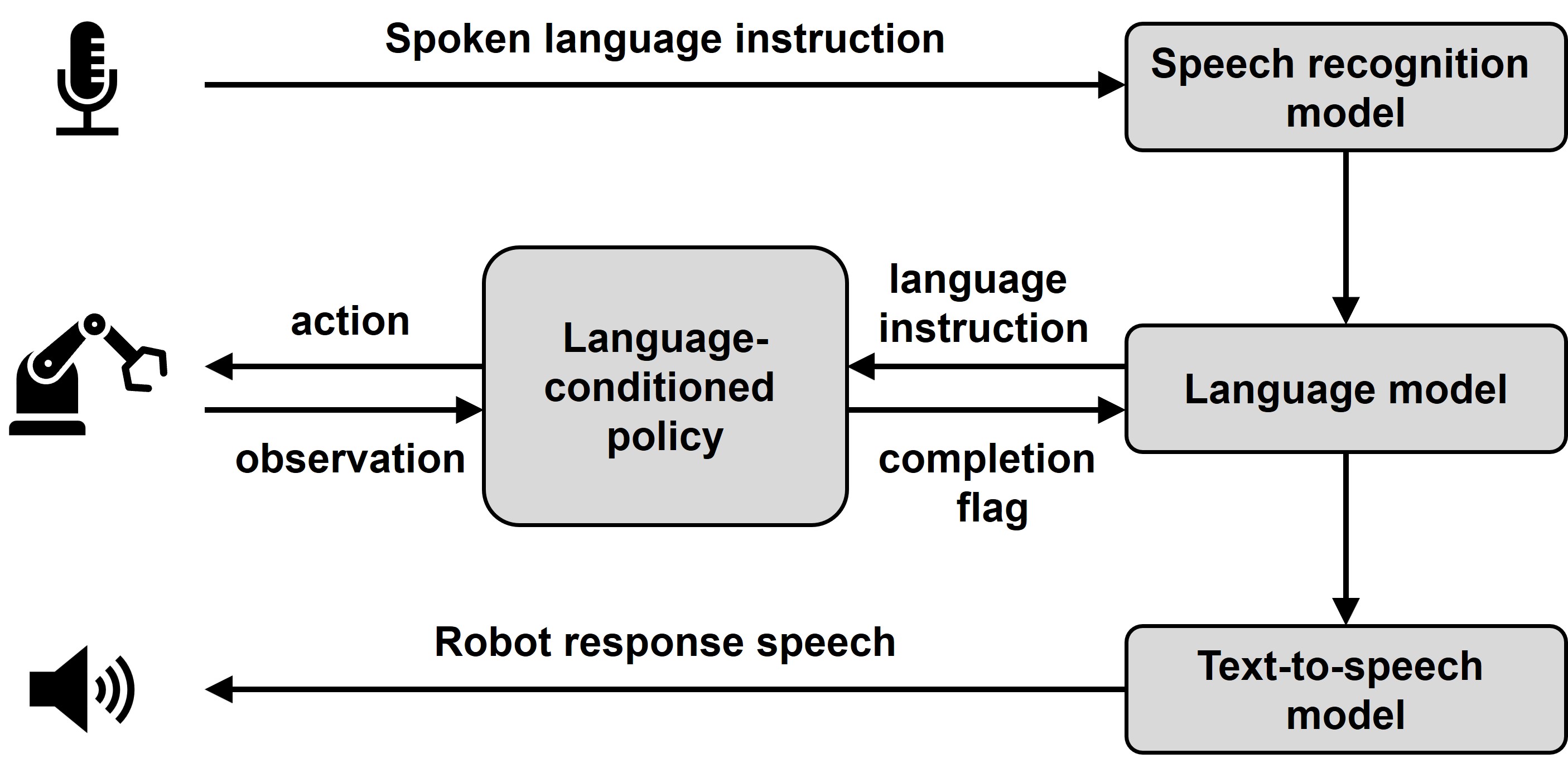}\\
  \begin{minipage}{0.9\columnwidth}
    \begin{center} \footnotesize (A) System flow \end{center}
  \end{minipage}\\
  \vspace{3mm}
  \includegraphics[width=0.9\columnwidth]{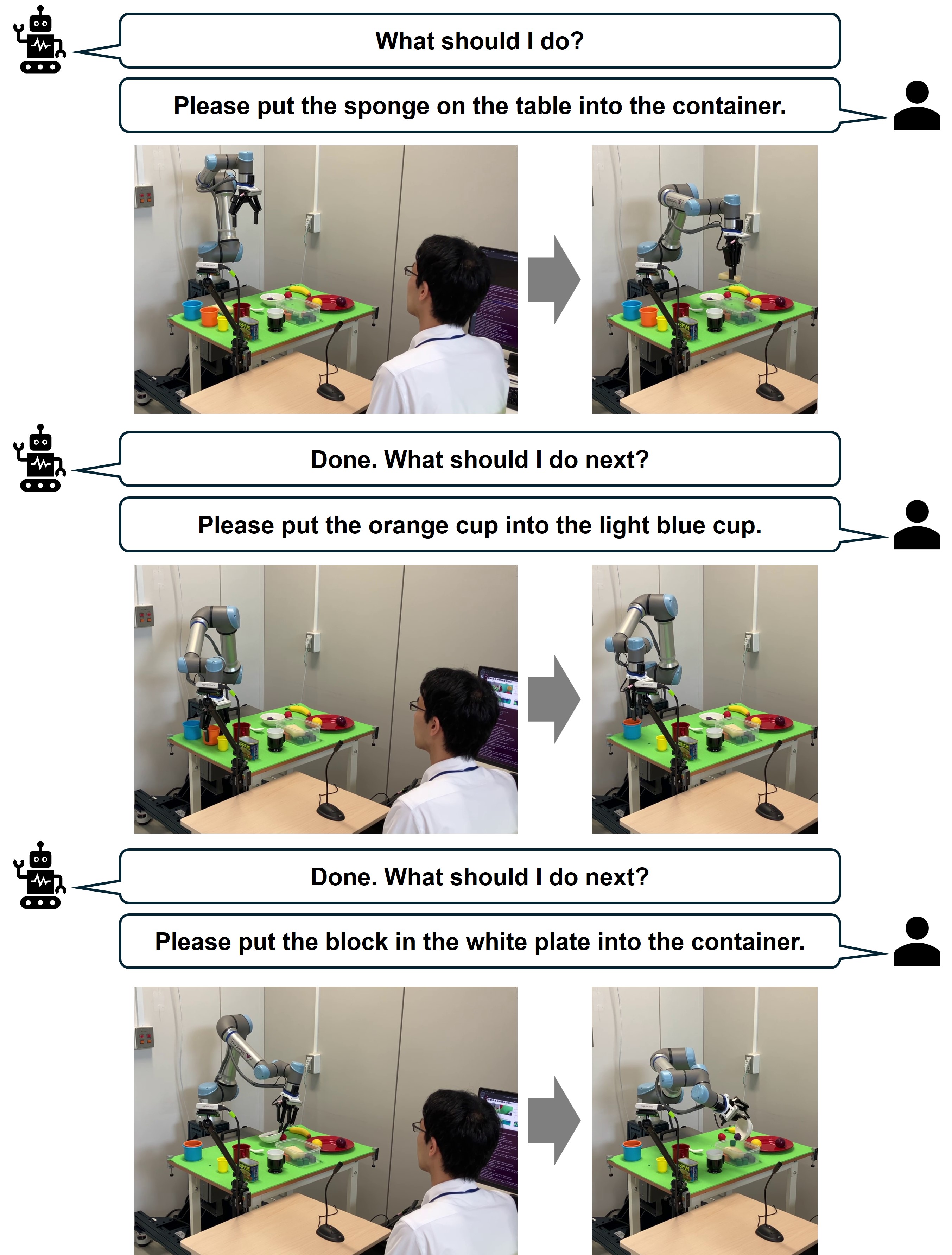}\\
  \begin{minipage}{0.9\columnwidth}
    \begin{center} \footnotesize (B) Example of interactive demonstration \end{center}
  \end{minipage}
  \caption{
    Interactive robotic system.\newline
    \footnotesize{By integrating speech recognition and language models, we constructed a manipulation system that receives instructions through natural language dialogue with a user. The demonstration shown here was conducted in Japanese, although the system can be easily adapted to other languages.}
  }
  \label{fig:interactive}
\end{figure}

We present an interactive robotic system integrating RoboManipBaselines with speech and language models.
We consider a scenario in which multiple objects are placed on a tabletop and a user provides instructions through spoken language.

As shown in \figref{fig:interactive}~(A), the human–robot dialogue is implemented through the following pipeline.
\begin{itemize}
  \item \textbf{Phase 1:} The user's speech is converted into text using a speech recognition model.
  \item \textbf{Phase 2:} A language model selects the most appropriate instruction from a predefined list based on the recognized text. If no suitable instruction is found, the system asks the user to repeat the command and returns to Phase~1.
  \item \textbf{Phase 3:} A language-conditioned policy is rolled out to execute the corresponding manipulation task.
  \item \textbf{Phase 4:} After task completion, the robot requests the next instruction using a text-to-speech model and the process returns to Phase~1.
\end{itemize}

The speech recognition, language, and text-to-speech components were implemented using the OpenAI API (e.g., Whisper and GPT-4o).
Because the dialogue processing (Phases~1, 2, and 4) and robot control (Phase~3) involve different models and execution frequencies, they were implemented as separate processes and connected via socket communication.
Only minimal changes to the rollout implementation of RoboManipBaselines were required to pass the received language instruction to the policy\footnote{The code is available at:\\\scriptsize{\url{https://github.com/isri-aist/RoboManipBaselines/tree/interactive-demo}}}.

As the language-conditioned policy, we used Multi-Task ACT (MT-ACT)~\cite{MtAct:Bharadhwaj:ICRA2024}, which can be trained and executed with relatively modest computational resources.
The system itself is not tied to a specific policy and can be easily replaced with other language-conditioned policies.
To train the MT-ACT policy, we defined 20 subtasks corresponding to combinations of objects and target locations (e.g., ``place the banana on the red plate'').
For each subtask, 50 demonstration episodes were collected, resulting in a dataset of 1000 episodes (approximately six hours of data).
Preliminary evaluation showed that the trained policy successfully completed more than half of the subtasks, with most failures occurring when grasping small objects.

Using this system, we verified that multiple subtasks can be executed sequentially through online voice interaction, as shown in \figref{fig:interactive}~(B).
Executing subtasks sequentially changes the object arrangement from the training scene, yet the policy was still able to perform subsequent tasks.
The current system selects instructions from the 20 predefined subtasks used during training.
Since the trained policy also demonstrated some generalization to language instructions not present in the training data, future work will extend the system to handle more open-ended language commands.

\subsection{Depth Camera Comparison for 3D Diffusion Policy}

\begin{figure}[tb]
  \centering
  \includegraphics[width=0.53\columnwidth]{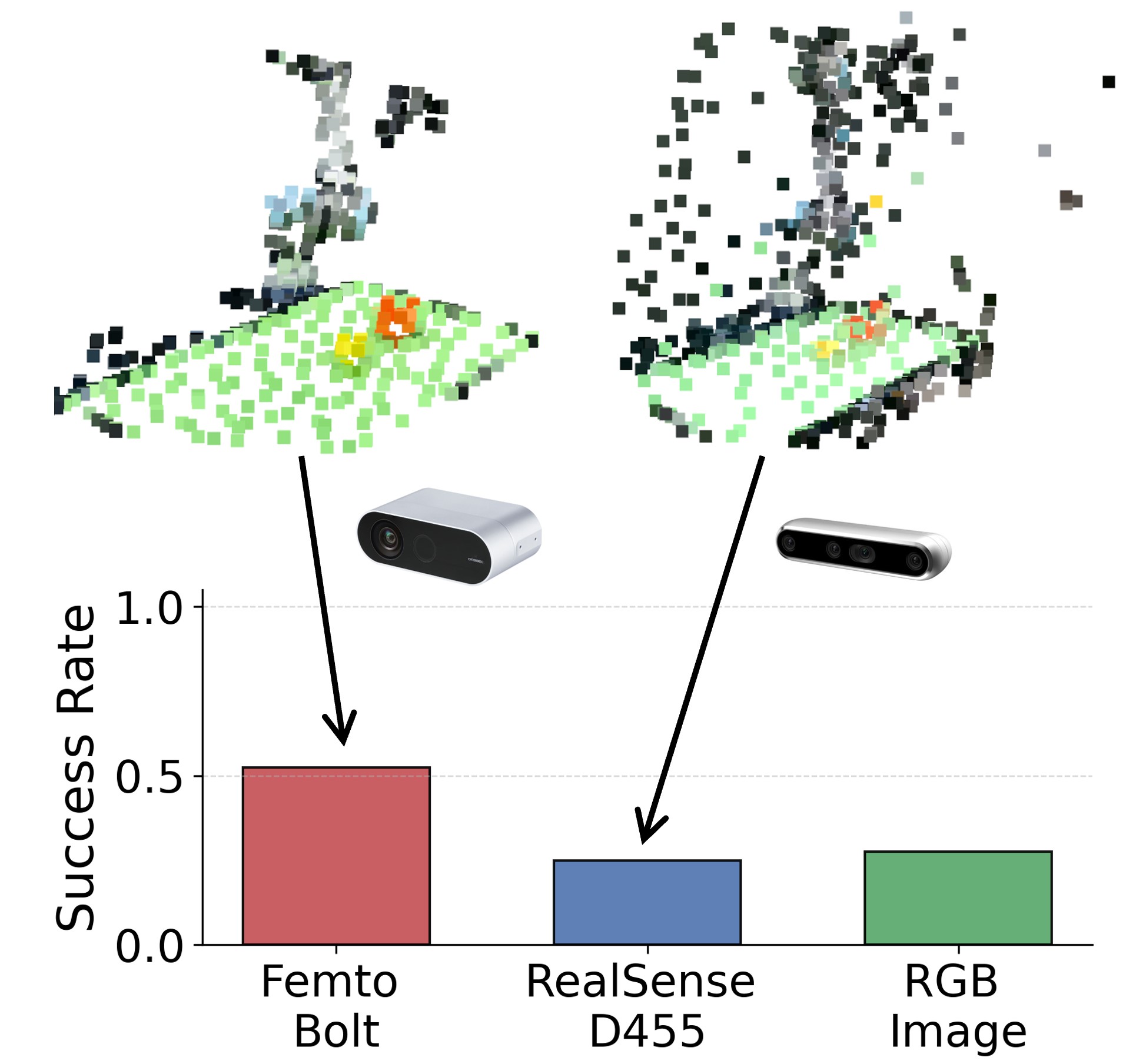}
  \includegraphics[width=0.45\columnwidth]{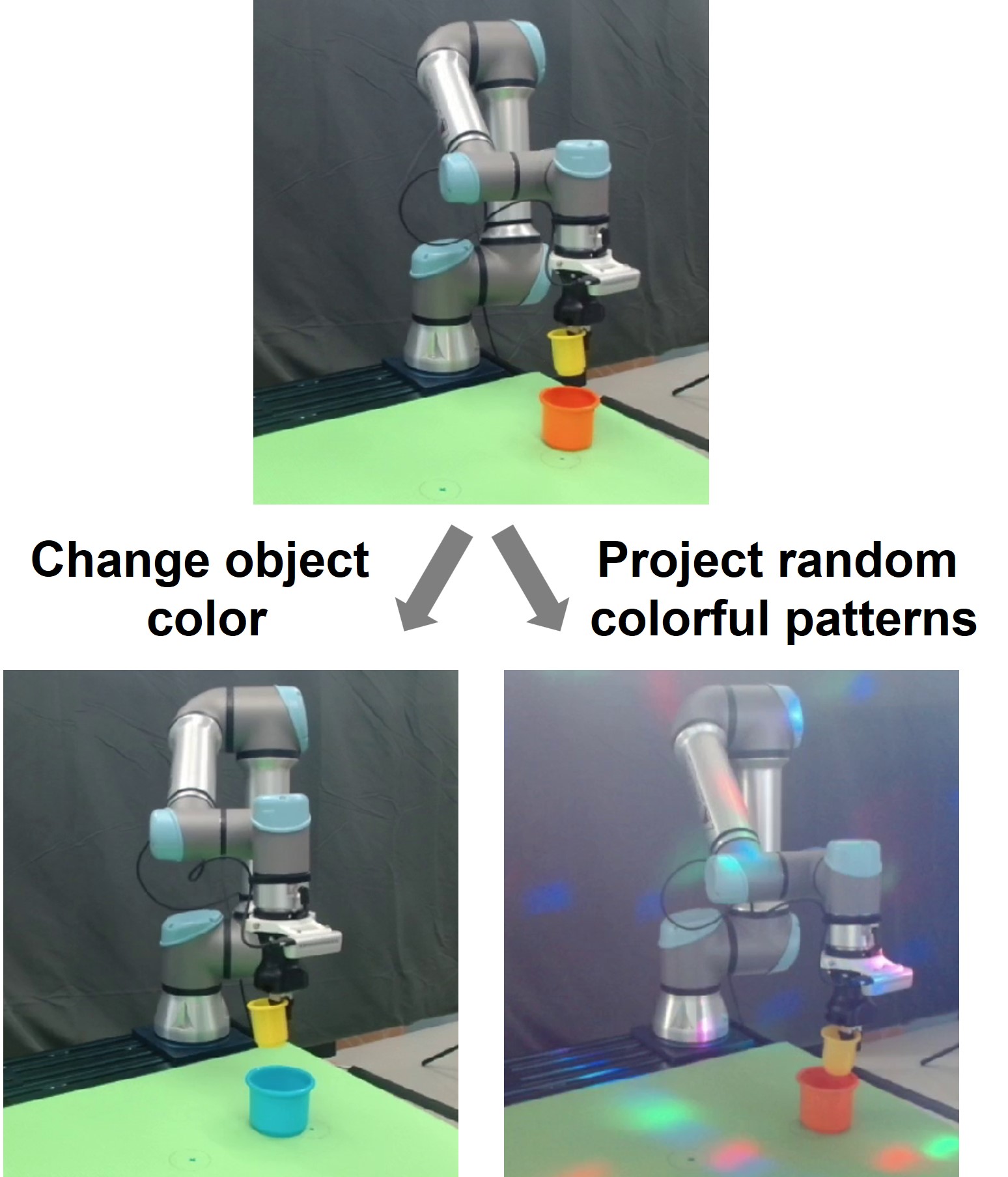}\\
  \begin{minipage}{0.53\columnwidth}
    \begin{center} \footnotesize (A) Experimental results \end{center}
  \end{minipage}
  \begin{minipage}{0.45\columnwidth}
    \begin{center} \footnotesize (B) Visual disturbances \end{center}
  \end{minipage}
  \caption{
    Depth camera comparison for 3D Diffusion Policy.\newline
    \footnotesize{We compare the performance of 3D Diffusion Policy using two depth cameras.
      The upper left shows the point cloud input to the policy after cropping and downsampling.
      The stereo-based RealSense D455 produces noticeable noise even in free space, whereas the ToF-based Femto Bolt yields cleaner point clouds.
      40 rollouts were performed for each setting, showing that point cloud quality strongly affects task success rates.
      In half of the rollouts, visual disturbances not present during training were introduced.
      While 3D Diffusion Policy using geometry-only point clouds remained robust, the RGB-based Diffusion Policy showed a significant performance drop.}
  }
  \label{fig:depth}
\end{figure}

We present a depth camera comparison for 3D Diffusion Policy~\cite{3DDiffusionPolicy:Ze:RSS2024} using RoboManipBaselines.
The framework supports multiple depth cameras for policies based on 3D point clouds.
\figref{fig:depth} shows the evaluation results using the stereo-based Intel RealSense D455 and the time-of-flight (ToF) camera ORBBEC Femto Bolt.
A clear performance difference is observed depending on the depth camera, with the ToF-based camera achieving higher task success rates.
Notably, the original 3D Diffusion Policy study~\cite{3DDiffusionPolicy:Ze:RSS2024} also used a ToF-based camera (RealSense L515), which is consistent with our observation.
In this experiment, visual disturbances were introduced by changing object colors and projecting random lighting patterns.
Since 3D Diffusion Policy uses colorless point clouds, it is unaffected by such visual disturbances, whereas the RGB-based Diffusion Policy shows a noticeable performance degradation.

RoboManipBaselines provides a unified interface for handling point clouds in both real-world and simulation environments.
Policies based on 3D point clouds typically require preprocessing such as converting depth images into point clouds, cropping, and downsampling.
The framework provides utility functions for these steps across multiple depth cameras, enabling straightforward deployment of 3D policies including 3D Diffusion Policy and ManiFlow Policy~\cite{ManiFlowPolicy:Yan:CoRL2025}.



\subsection{Hardware Extension for Contact Sensing}

\begin{figure}[tb]
  \centering
  \includegraphics[width=0.35\columnwidth]{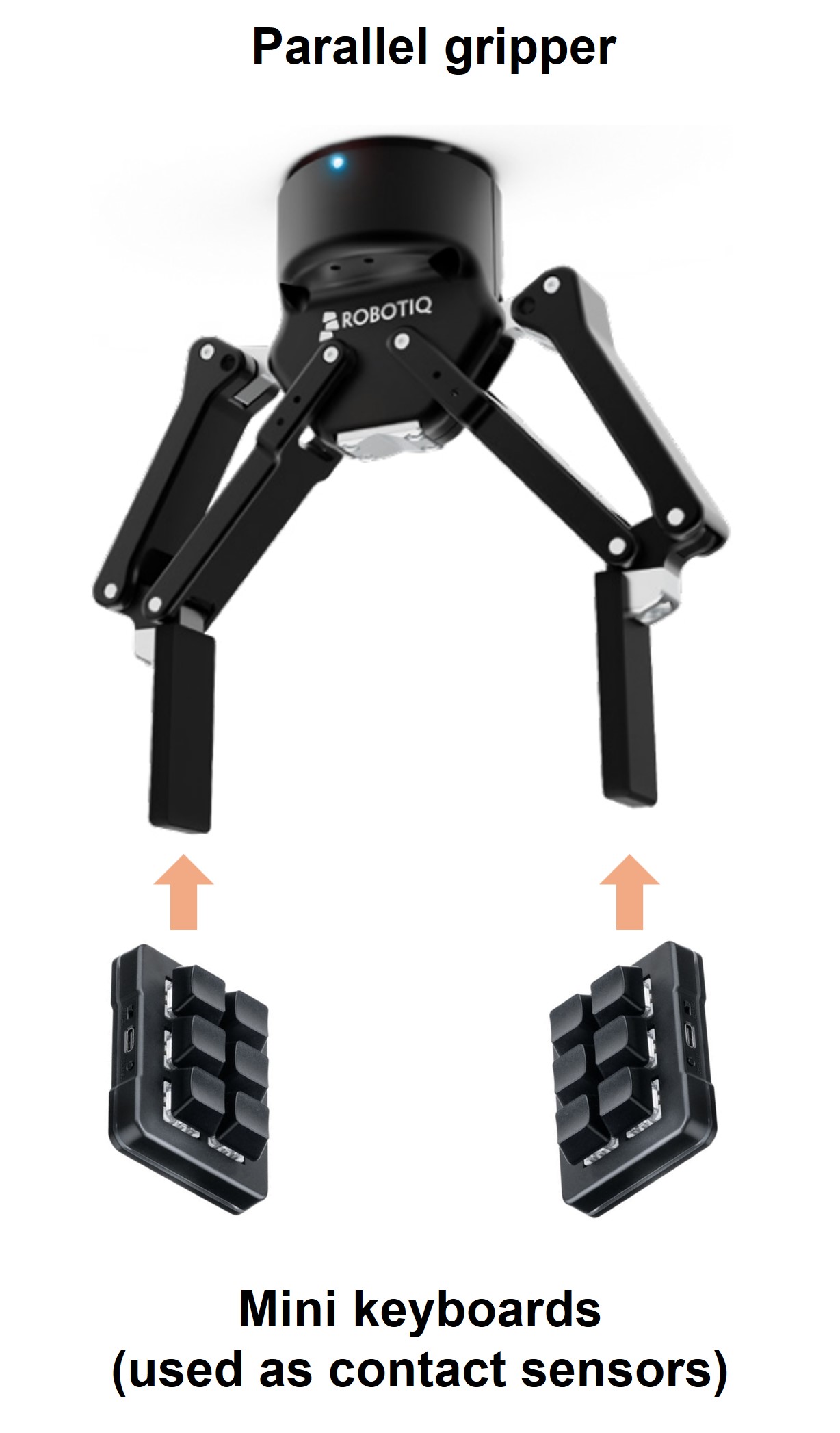}
  \includegraphics[width=0.63\columnwidth]{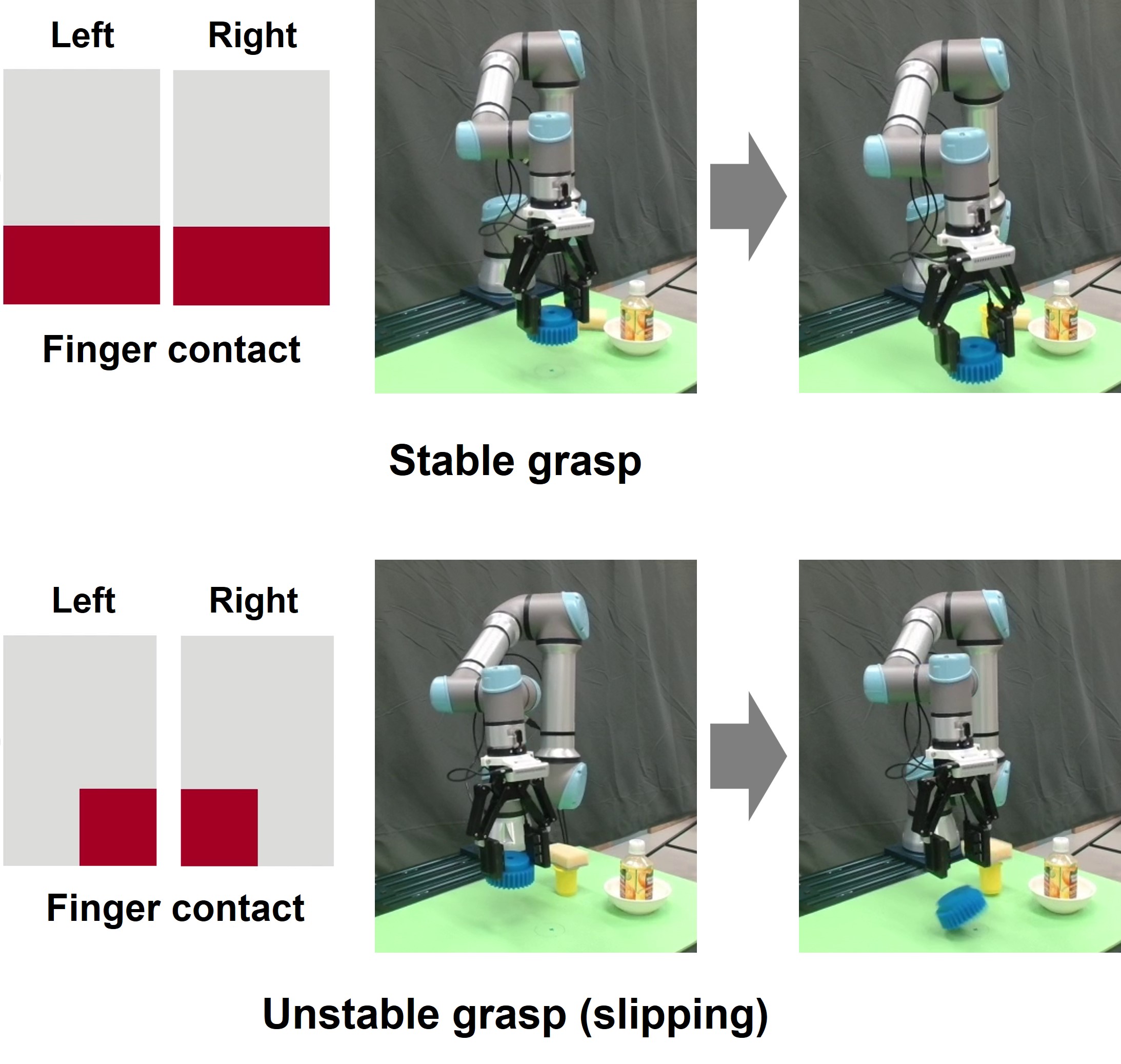}\\
  \begin{minipage}{0.35\columnwidth}
    \begin{center} \footnotesize (A) Hardware \end{center}
  \end{minipage}
  \begin{minipage}{0.63\columnwidth}
    \begin{center} \footnotesize (B) Teleoperation \end{center}
  \end{minipage}\\
  \vspace{4mm}
  \includegraphics[width=0.98\columnwidth]{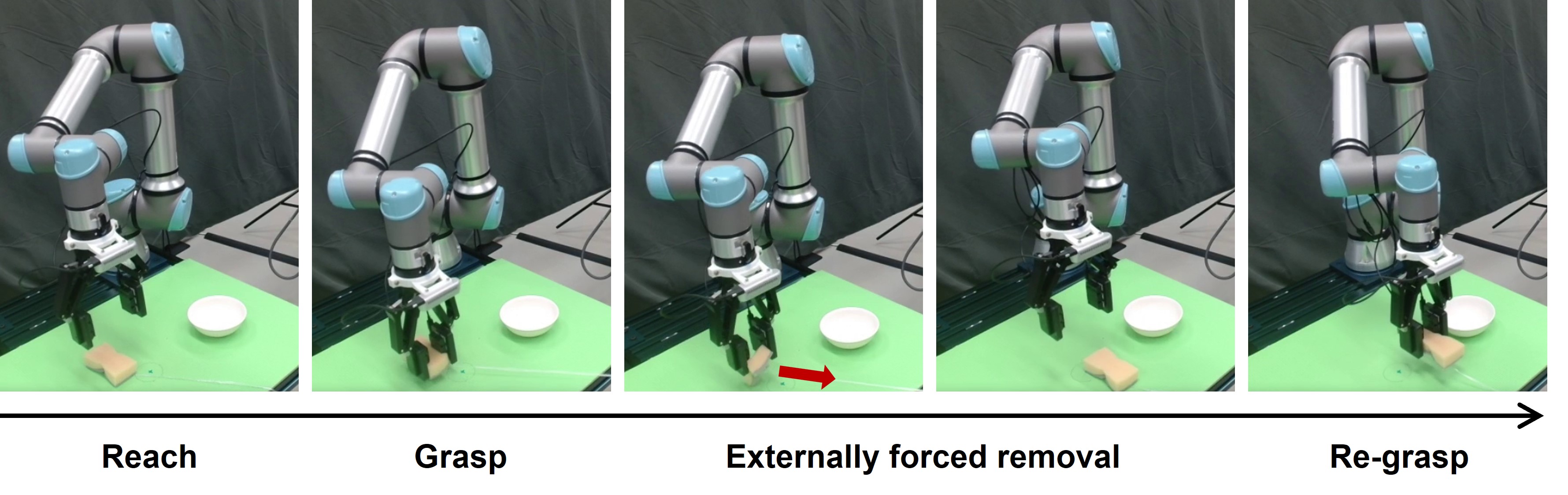}\\
  \begin{minipage}{0.9\columnwidth}
    \begin{center} \footnotesize (C) Policy rollout with contact sensing \end{center}
  \end{minipage}
  \caption{
    Hardware extension for contact sensing.\newline
    \footnotesize{A compact keyboard with $2 \times 3$ keys is mounted on the fingers of a parallel gripper and used as a grid contact sensor for imitation learning.}
  }
  \label{fig:keyboard}
\end{figure}

We present a prototype hardware extension that uses a compact keyboard as a contact sensor with RoboManipBaselines.
Contact information inside a gripper is important for grasping~\cite{3D-ViTac:Huang:CoRL2024}, but typical tactile sensors are often expensive, fragile, or require dedicated interfaces.
To explore a simpler alternative, we mounted a compact keyboard with $2 \times 3$ keys on the two fingers of a parallel gripper and used the key press states as a grid-based contact measurement, as shown in \figref{fig:keyboard}~(A)\footnote{A Robotiq 2F-140 parallel gripper was used. The keyboard used in this prototype was the Sanwa Direct 400-SKB081 programmable keyboard, although other similar devices can be used. \\\scriptsize{\url{https://direct.sanwa.co.jp/ItemPage/400-SKB081}}}.
Since such keyboards are low-cost, widely available, and use a standard USB interface, they provide a practical alternative for exploring contact sensing in manipulation learning.

In RoboManipBaselines, this functionality was implemented simply by adding the keyboard key states as contact observations to the real-world environment.
This enables teleoperation-based data collection with contact sensing, as well as policy training and rollout using the same information\footnote{This functionality has been integrated into the RoboManipBaselines repository:\\\scriptsize{\url{https://github.com/isri-aist/RoboManipBaselines}}}.
This is possible because data collection, policy training, and rollout use a unified environment interface compatible with OpenAI Gym~\cite{OpenAIGym:Brockman:arXiv2016}.

As shown in \figref{fig:keyboard}~(B), different grid contact patterns were observed during teleoperation depending on grasp stability.
Furthermore, when a Diffusion Policy~\cite{DiffusionPolicy:Chi:IJRR2024} was trained with this contact information and deployed, the robot attempted to re-grasp the object when it was forcibly removed during manipulation, as illustrated in \figref{fig:keyboard}~(C).
This behavior suggests that the policy used contact measurements to determine whether an object was in the gripper.
These results demonstrate that RoboManipBaselines enables rapid prototype studies of imitation learning systems, including hardware extensions.




\section{Limitations}

This framework has several limitations. First, training imitation learning policies may require substantial GPU resources and memory, and the computational cost can be non-negligible. In addition, the current implementation depends on a Linux environment and a Python-based software stack, and support for other operating systems and programming languages is limited.

Second, support for real-world robots is currently limited, and the environments evaluated in this paper are restricted to a subset of manipulators. Therefore, applying the framework to other robotic hardware may require minimal additional implementation.

Furthermore, this framework is primarily designed for imitation learning in robotic manipulation and is not necessarily suited for workflows centered on mobile robot navigation or reinforcement learning. Applying it to such tasks would require additional design and extensions.

In addition, the benchmark results presented in this paper exhibit relatively high variance across random seeds, and the number of trials in real-world experiments is limited. Therefore, the reported success rates should be interpreted with caution, and the relative performance differences between policies may not always be statistically significant.

Moreover, although the framework supports both simulation and real-world datasets within a unified interface, this paper does not explore training with mixed datasets. Techniques such as domain randomization and sim-to-real transfer are not explicitly addressed, and investigating how to combine these data sources remains an important direction for future work.

Finally, although this paper mentions support for humanoid robots and mobile manipulators, the current implementation treats humanoids by fixing the legs and using only the upper body. As a result, whole-body tasks involving locomotion are not directly supported, and extending to more general settings remains future work.





\section{Conclusion}

In this paper, we introduced RoboManipBaselines, a unified framework for imitation learning in robotic manipulation across both simulation and real-world environments.
The framework systematically implements the core components of imitation learning, namely the environment, dataset, and policy, while emphasizing integration, generality, extensibility, and reproducibility.
This design enables the entire imitation learning pipeline, from data collection to policy training and rollout, to be executed in a unified manner across both simulation and real-world environments, independent of the robot platform or policy type.

We also conducted benchmark evaluations on both simulation and real-world manipulation tasks using the framework and presented comparative results across multiple policies.
Furthermore, we introduced several research applications, including data augmentation, integration with tactile models, interactive robotic systems, depth camera evaluation, and hardware extensions.
These examples demonstrate that RoboManipBaselines facilitates the implementation and validation of new research ideas.

The user community of the framework is also expanding.
Through talks, hands-on seminars, and technical collaborations with industry, several companies have evaluated the framework on their proprietary robots.

While RoboManipBaselines can be used as a standalone framework, it can also be used in a complementary manner with other existing frameworks through data format conversion and connections to simulation interfaces.
In future work, we plan to further strengthen interoperability with other frameworks and continue extending the system to support a wider range of robotic manipulation research.
We hope that RoboManipBaselines will contribute to accelerating research and experimentation in robotic manipulation.





\bibliographystyle{IEEEtran}
\bibliography{main.bib}

\end{document}